\documentclass{article}

\usepackage[preprint]{neurips_2026}

\usepackage[utf8]{inputenc} 
\usepackage[T1]{fontenc}    
\usepackage{hyperref}       
\usepackage{url}            
\usepackage{booktabs}       
\usepackage{amsfonts}       
\usepackage{nicefrac}       
\usepackage{microtype}      
\usepackage{xcolor}         
\usepackage{amsmath}
\usepackage{graphicx}
\usepackage{subcaption}
\usepackage{wrapfig}
\usepackage{tikz}
\usetikzlibrary{arrows, matrix}
\usepackage{colortbl}
\usepackage{enumitem}
\usepackage{multirow}

\usepackage{xspace}

\newcommand{\DIMS}{\textsc{DiMS}\xspace}

\usepackage{amsthm}
\newtheorem{theorem}{Theorem}
\newtheorem{lemma}{Lemma}
\newtheorem{remark}{Remark}

\newcommand{\RR}{\mathbb{R}} 
\newcommand{\btheta}{\boldsymbol{\theta}}
\newcommand{\balpha}{\boldsymbol{\alpha}}

\newcommand{\bx}{\boldsymbol{x}}
\newcommand{\bv}{\boldsymbol{v}}
\newcommand{\tbv}{\tilde{\boldsymbol{v}}}
\newcommand{\bp}{\boldsymbol{p}}
\newcommand{\bgamma}{\boldsymbol{\gamma}}

\newcommand{\beps}{\boldsymbol{\epsilon}}
\newcommand{\bmu}{\boldsymbol{\mu}}
\newcommand{\M}{\mathcal{M}}
\newcommand{\TM}[1]{\mathcal{T}_{#1}\mathcal{M}}
\newcommand{\I}{\mathbb{I}}
\newcommand{\J}[2]{\mathbf{J}_{#1}(#2)}
\newcommand{\G}[1]{\mathbf{G}(#1)}
\newcommand{\invG}[1]{\mathbf{G}(#1)^{-1}}
\newcommand{\sqrtinvG}[1]{\mathbf{G}(#1)^{-\frac{1}{2}}}
\newcommand{\grad}[2]{\nabla_{#1}#2}
\newcommand{\Loss}[1]{\mathcal{L}(#1)}

\definecolor{lightgray}{gray}{0.9}

\DeclareRobustCommand{\tealline}{\tikz[baseline=-0.6ex]\draw[teal, line width=2pt, line cap=round] (0,0) -- (1em,0);}
\DeclareRobustCommand{\orangeline}{\tikz[baseline=-0.6ex]\draw[orange, line width=2pt, line cap=round] (0,0) -- (1em,0);}

\hypersetup{
    colorlinks,
    linkcolor={red!50!black},
    citecolor={blue!50!black},
    urlcolor={blue!80!black}
}

\title{Don't Stop Me \emph{Yet}: Sampling Loss Minima via Dissipative Riemannian Mechanics}

\author{%
    Albert Kj{\o}ller Jacobsen$^{1}$ 
    \And
    Leo Uhre Jakobsen$^{2}$ 
    \And
    Johanna Marie Gegenfurtner$^{1}$ 
    \And 
    Georgios Arvanitidis$^{1}$ \\  \\[-5pt]
    $^{1}$ Section for Cognitive Systems, Technical University of Denmark \\
    $^{2}$ Center for Quantum Information Physics, New York University \\
    \texttt{\{akjja, johge, gear\}@dtu.dk}, \quad \texttt{luj2004@nyu.edu}
}

\begin{document}

\maketitle

\vspace{-12pt}
\begin{abstract}
\vspace{-5pt}
The minima of modern neural network loss functions are typically not isolated, rather they form connected components of reparameterization invariant solutions on the training data. 
Analytically characterizing these solutions is a hard problem, but sampling approaches are feasible. 
By construction, existing methods either spread over low-loss regions, and thus do not sample reparameterization invariant solutions exactly, or are inherently local, which limits exploration of other minima valleys.
We propose sampling such reparameterization invariant models using a dynamical system based on kinetic energy, subject to a gravitational pull and a friction term that dissipates energy from the system. 
Our proposed sampler, \textsc{DiMS}, is guaranteed to sample exactly from the minimum level sets and depends on physically motivated hyperparameters which allows control over the exploration capabilities of the sampler.
We consider uncertainty quantification in Bayesian inference as the motivating problem and observe improved performance compared to previously proposed approaches.
\end{abstract}

\vspace{-9pt}
\section{Introduction}
\vspace{-3pt}
\begin{wrapfigure}{r}{0.37\textwidth}
    \centering
    \vspace{-30pt}
    \includegraphics[width=0.7\linewidth]{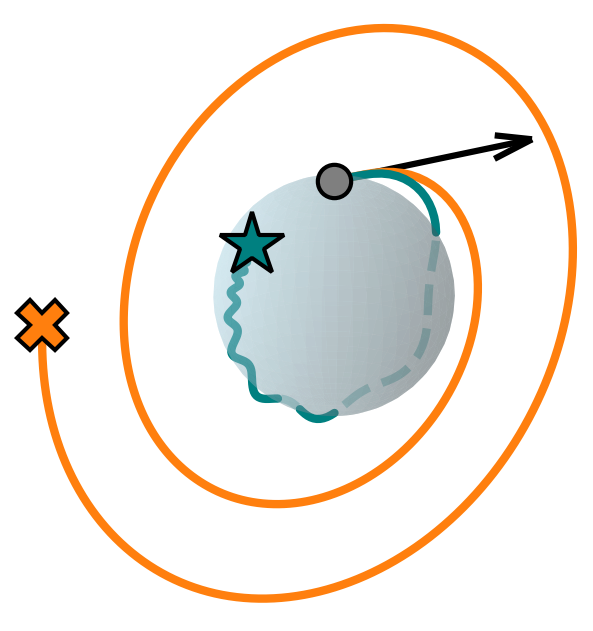}
    \caption{Two trajectories through parameter space via different dynamical systems. The sphere constitutes a submanifold of parameters that minimize $\mathcal{L}(\btheta)=\sum_{i=1}^K \theta_i^2-1$. While a geodesic on the loss surface (\orangeline) is the straightest path, it never converges to a minimum solution, even when starting from one. Our modified dynamics (\tealline) ensure energy dissipation which leads the particle to stop at some minimum solution. 
    }
    \label{fig:figure1}
    \vspace{-15pt}
\end{wrapfigure}

Training a deep neural network results in finding a single point estimate with minimum loss within a vast, high-dimensional parameter space. Yet the loss landscape has a rich geometric structure and a single point estimate can only capture local information about it. This can have important shortcomings for downstream tasks, for instance, in Bayesian inference. The architecture and the data together shape the loss landscape such that several parameter configurations perform equally well on the training data; the minimum level sets of modern overparameterized networks form extended, continuous submanifolds for which the model is \emph{reparameterization invariant} on the observed data \citep{pmlr-v70-dinh17b, visualizing2018li, modeconnectivity2018garipov, pmlr-v80-draxler18a}. 

In this paper we ask \emph{``how should we navigate the space of good solutions''} rather than  \emph{``where is a good solution''}. We propose the \textbf{Di}ssipative \textbf{M}inima \textbf{S}ampler (\DIMS), formulated as a dynamical system that navigates the loss landscape into low-loss regions while guaranteeing convergence to the submanifold of minimum solutions. As a result, the gathered samples correspond to a diverse set of functions that agree on the training data but vary in their predictions elsewhere. 

Recent works have focused on sampling low-loss regions by respecting the inherent geometry of the problem \citep{bergamin2023riemannian, yu2024riemannian, fadel2025viking}, and even restricting samples to the reparameterization invariant set \citep{roy2024reparameterization}. Our proposed method differ by guaranteeing convergence to the minimum manifold while allowing for escaping the initial loss basin to find other connected minima - a combination that neither optimization-based methods nor existing sampling-based methods are designed for. Our contributions include:
\begin{enumerate}[leftmargin=*, itemsep=1pt, topsep=1pt]
    \item \textbf{A deterministic sampler for the minimum submanifolds}, continuously visiting distinct points on the minimum submanifolds of the training loss. We formalize geometry-driven dynamics on Riemannian manifolds induced by the loss with physically interpretable hyperparameters controlling the exploration behavior of the sampler. The resulting trajectories can be considered as an implicit approximate Bayesian posterior concentrated on the minimum manifold.
    \item \textbf{Theoretical guarantees} that our dynamics explore the loss surface before converging to the minimum manifold in finite time.
    \item \textbf{Empirical validation} that posterior approximations concentrated on the minimum manifold generalize well, and can improve predictive performance, calibration, and especially out-of-domain uncertainty in Bayesian inference settings. 
\end{enumerate}

\begin{figure}[tb]
    \centering
    \vspace{-9pt}
    \includegraphics[width=1.0\linewidth]{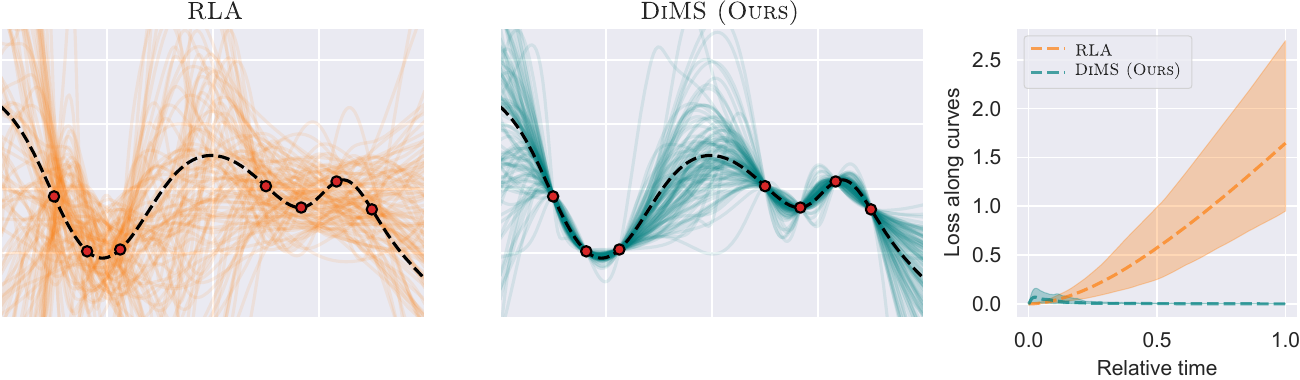}
    \vspace{-13pt}
    \caption{We fit a neural network to $N=7$ training points and show function-space samples from two geometry-aware sampling schemes formulated as continuous dynamical systems. The Riemannian Laplace approximation (\textit{left}) explores low-loss regions but does not enforce interpolating the training data. Our sampler \textsc{DiMS} (\textit{center}) leverages an alternative formulation of the system and guarantees convergence to minimum training loss solutions, ensuring all function-space samples to interpolate the training data, while remaining in low-loss regions along paths generated by the dynamics (\textit{right}).
    }
    \vspace{-7pt}
    \label{fig:sinusoidal}
\end{figure}

\paragraph{Intuition:} 
The sampling process (Figure \ref{fig:figure1}) starts by shooting out a particle from a fixed point $\btheta^\ast$ that minimizes the loss. The direction and initial speed of the particle is randomly sampled from a distribution, $\bv \sim q(\bv)$, and determines the initial kinetic energy of the particle. By introducing external forces well-known from physics, we guide the particle toward minima, eventually stopping at a minimum level set. Every intersection with this set corresponds to a model that optimally fits the training data, yet exhibits different behavior in other data regions. This leads to high functional variation in regions away from observed data (Figure \ref{fig:sinusoidal}), which is desired in Bayesian inference settings. Remark that \DIMS is not a traditional approximate posterior as it deliberately generates samples only at the maxima ridges of the true posterior.

\section{Background - geometry of loss surfaces}
Let $\mathcal{D} = \{(\bx_i, y_i)\}_{i=1}^N$ denote a dataset with $N$ paired input-output observations of $\bx_i \in \mathcal{X} \subseteq \RR^D$ and $y_i \in \mathcal{Y}$ where $\mathcal{Y}\subseteq\RR$ for regression or $\mathcal{Y}=\{1, \dots, C\}$ for classification among $C$ classes. In most predictive settings we can define a likelihood parameterized by a neural network, then following Bayes' theorem the negative log-posterior equals the log-joint up to a constant:
\begin{equation}
    \label{eq:log-posterior-loss}
    \Loss{\btheta; \mathcal{D}} := -\log p(\mathcal{D}|\btheta) -\log p(\btheta) + \log p(\mathcal{D}) = -\log p(\btheta) - \sum\nolimits_{i=1}^N\log p\left(y_i| f_{\btheta}(\bx_i)\right) + C
\end{equation}
where $f_{\btheta}: \mathcal{X} \rightarrow \mathcal{Y}$ is the neural network with parameters $\btheta \in \Theta$ and $p(\btheta)$ is a prior distribution over the parameter space $\Theta \subseteq \RR^K$. The posterior distribution $p(\btheta|\mathcal{D})$ is intractable when working with neural networks and we thus resort to an approximate posterior $q(\btheta)$ rather than the true one. Using the generalized Bayesian framework, we formulate a Gibbs posterior over the parameter space that enables a Bayesian treatment of any neural network without requiring a likelihood specification.
\begin{figure}[tb]
    \centering
    \vspace{-10pt}
    \includegraphics[width=0.9\linewidth]{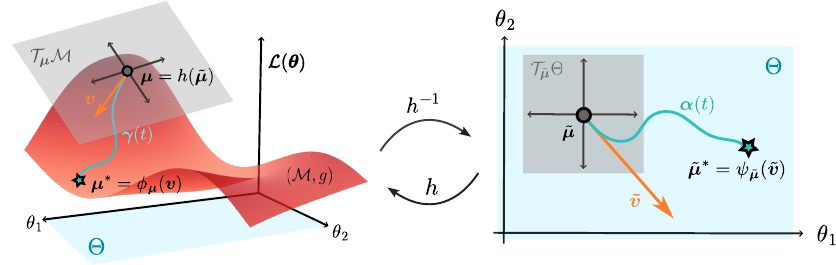}
    \caption{The loss surface viewed as a Riemannian manifold $(\M, g)$ embedded in $\RR^{K+1}$. The parameter space inherits the geometry of the loss surface via the pullback metric $\G{\btheta}$. A curve on the manifold $\bgamma_{\bv}(t) \in \M$ starting at $\bmu$ is generated by the initial velocity $\bv$ and can be expressed through its corresponding path in the parameter space $\balpha_{\tbv}(t) \in \Theta$ induced by $\tbv$. We denote the mappings from the initial velocities to the curve at time $t$ by $\phi_{\bmu,t}:\TM{\bmu}\rightarrow \M$ and $\psi_{\tilde{\bmu}, t}:\mathcal{T}_{\tilde{\bmu}}\Theta \rightarrow \Theta$. 
    }
    \vspace{-5pt}
    \label{fig:curve-intuition}
\end{figure}

We briefly review Riemannian geometry of loss surfaces, which forms the foundation of our approach. For notational convenience we suppress the dependency on the data $\mathcal{D}$ unless explicitly needed.

\subsection{Defining the loss surface as a Riemannian manifold}

Given a smooth loss function $\mathcal{L}: \RR^K \rightarrow \RR$, the graph of the function over the parameter space defines a smooth, $K$-dimensional submanifold embedded in $\RR^{K+1}$, often called the loss surface:
\begin{equation}
    \label{eq:embedding}
    \M=\left\{h(\btheta) \mid \btheta \in \Theta\right\} \subseteq \RR^{K+1}, \qquad \textrm{with}\qquad  h(\btheta) = \left(\theta_1, \dots, \theta_K, \mathcal{L}(\btheta)\right).
\end{equation}
This definition of $\M$ naturally induces a Riemannian metric on the parameter space, as we now establish. For a configuration $\btheta$, the ambient tangent space centered at $h(\btheta)$ is spanned by the Jacobian
$\J{h}{\btheta} = \grad{\btheta}{h(\btheta)}
    = \begin{bmatrix} 
        \I_K, \grad{\btheta}{\Loss{\btheta}}
    \end{bmatrix}^\top
    \in \RR^{(K+1) \times K}
$, and we can express two tangent space vectors $\bv_1, \bv_2 \in \TM{\bp}$ in parameter space coordinates as $\bv_1=\mathbf{J}_h(\btheta) \tbv_1$ and $\bv_2=\mathbf{J}_h(\btheta)\tbv_2$ where $\tbv_1, \tbv_2 \in \TM{\btheta}$ denote the corresponding tangent vectors in parameter space coordinates. We can then define inner products on the manifold via the metric tensor $g_{\bp}: \TM{\bp} \times \TM{\bp} \rightarrow \RR$ as:
\begin{equation}
    g_{\bp}(\bv_1, \bv_2) = \bv_1^\top \bv_2 = \tbv_1^\top \mathbf{J}_h(\btheta)^\top \mathbf{J}_h(\btheta) \tbv_2 = \tbv_1^\top \mathbf{G}(\btheta) \tbv_2.
\end{equation}
Here $\G{\btheta} = \I_K + \grad{\btheta}{\Loss{\btheta}} \grad{\btheta}{\Loss{\btheta}}^\top$ is the symmetric positive-definite (SPD) metric matrix, making $(\M, g)$ a Riemannian manifold whose inner products are locally distorted by curvature information from the loss gradient. This definition of the inner product is the \emph{pullback metric} as it pulls the Euclidean inner product on the tangent space $\TM{\bp}$ to the parameter space $\Theta$, in which all geometric computations on $\M$ can therefore be carried out. Note that while the embedding $h$ from Equation \eqref{eq:embedding} captures only the loss geometry, the theory extends to any smooth embedding function.

\subsection{Loss-minimizing submanifolds}
For a non-convex, data-dependent loss function $\mathcal{L}:\Theta\rightarrow\RR$, we define the loss level set at $l$ as
$
    S_l(\mathcal{D}) = \bigsqcup_{k \in \mathcal{K}} \left\{\btheta \in \mathcal{B}_k: \Loss{\btheta} = l \geq \mathcal{L}^\ast(\mathcal{D})\right\},
$
where $\mathcal{B}_k$ is the basin of attraction of the $k$-th local minimum, $\mathcal{K}$ is the index set over all local minima and $\mathcal{L}^\ast(\mathcal{D})$ is the global minimum loss on data $\mathcal{D}$. 
We define the \emph{loss-minimizing submanifold} as the disjoint union of minimum level sets within basins:
\begin{equation}
    S_\ast(\mathcal{D}) = \bigsqcup\nolimits_{k\in \mathcal{K}} \left\{ \btheta \in \mathcal{B}_k: \Loss{\btheta} = \mathcal{L}_k^\ast(\mathcal{D}) \right\}.
\end{equation}
Here $\mathcal{L}^\ast_k(\mathcal{D})=\min_{\btheta^\prime\in \mathcal{B}_k} \Loss{\btheta^\prime; \mathcal{D}}$ denotes the minimum loss value of the $k$-th basin, so $S_{\ast}(\mathcal{D})$ is essentially the collection of all local minima.
A special case of the loss level submanifold is $S_{\mathcal{L}^\ast(\mathcal{D})}$ for which all solutions have minimal loss and are functionally similar on all points in $\mathcal{D}$. In the interpolation regime where $\mathcal{L}^\ast(\mathcal{D})=0$, all parameters in $S_0$ exactly interpolate the training data $\mathcal{D}$, yet may induce functionally distinct predictions on unseen inputs, as depicted in Figure \ref{fig:sinusoidal}.
We remark that in the overparameterized regime where $K \gg N$, the loss imposes at most $N$ constraints on the $K$-dimensional parameter space. With far more parameters than data points, there are locally at least $K-N$ directions in which the parameters can move without changing the loss value.

\section{Geometric mechanics - curves on manifolds induced by classical mechanics}
Classical mechanics is fundamentally the study of motion and thus naturally describes the evolution of a dynamical system over time. In the following, we recall the Lagrangian formulation for a single-particle system evolving in the intrinsic coordinates of a Riemannian manifold, which in our case is the parameter space $\Theta$. The trajectories will serve as the foundation for our sampler. 

The path of a particle traveling on the loss manifold $\bgamma:[t_0,t_1]\rightarrow \M$ can also be expressed as a parameter space curve $\balpha: [t_0, t_1]\rightarrow\Theta$, such that $\bgamma(t)=h(\balpha(t))$, illustrated in Figure \ref{fig:curve-intuition}. At any time along the path, we characterize the energy of the particle by its kinetic and potential energies:
\begin{equation}
    \label{eq:energies}
    E_{\text{kin}} = T(\balpha, \dot{\balpha}) =\frac{1}{2} \dot{\balpha}^\top \G{\balpha}\dot{\balpha} \qquad \qquad  
    E_{\text{pot}} = V(\balpha) = \kappa \cdot \left(\Loss{\balpha} - \Loss{\btheta^\ast_{\text{glob}}}\right)
\end{equation}
where $\dot{\balpha}=\frac{d}{dt}\balpha(t)$ is the velocity, and we follow standard notation conventions and neglect the explicit time-dependency unless needed. Note that $\btheta^\ast_{\text{glob}}$ denotes any global minimum and that $\kappa>0$ is a constant encoding the coupling of mass (which we set to unity) and gravitational acceleration. The particle dynamics are governed by the Euler-Lagrange (EL) equations, which are derived from the Lagrangian defined as the difference between kinetic and potential energy.
The EL equations simplify under the definition of the manifold (see Appendix \ref{appendix:geodesics}), and we get a second-order ordinary differential equation (ODE) governing the acceleration:
\begin{equation}
    \label{eq:acceleration-euler-lagrange}
    \ddot{\balpha} = - \left(\dot{\balpha}^\top \mathbf{H}_{\mathcal{L}}(\balpha)\dot{\balpha} + \kappa\right) \cdot \operatorname{grad}\Loss{\balpha}.
\end{equation}
where $\mathbf{H}_{\mathcal{L}}(\balpha):=\nabla_{\btheta}^2\Loss{\balpha}$ is the Hessian of the loss and $\operatorname{grad} \Loss{\balpha} = \invG{\balpha} \grad{\theta}{\Loss{\balpha}}$ is the Riemannian gradient, hence guiding motion toward the steepest descent, which corresponds to a gravitational pull on the particle traveling on the manifold. Remark that the Hessian term relates to the kinetic energy and that the $\kappa$ term relates to the potential energy. In practice, we can efficiently solve the second-order ODE numerically using automatic differentiation. 
\begin{wrapfigure}{r}{0.45\textwidth}
    \centering
    \vspace{-10pt}
    \includegraphics[width=1.0\linewidth]{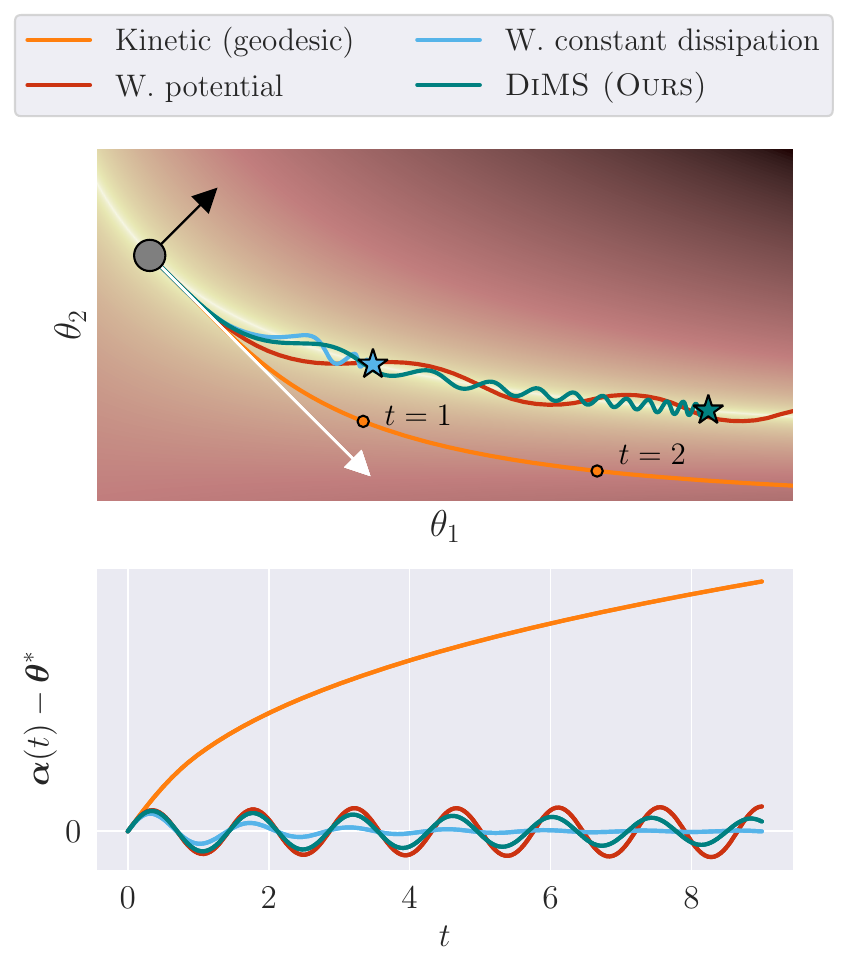}
    \caption{\textit{Top:} With the white initial velocity, the systems affected by a gravity pull remain closer to the minimum level set, yet without dissipation the particle never converges. With dissipation via a friction force, convergence is guaranteed, and our speed-dependent dissipation function allows traveling further than constant friction. \textit{Bottom:} With the black initial velocity, curves affected by gravity oscillates around the minimum, while the dissipative systems converge steadily at different speeds. The free particle can travel forever.}
    \label{fig:example-2D}
    \vspace{-50pt}
\end{wrapfigure}

In the following two paragraphs, we highlight key properties, and the main issue with the dynamical system developed so far for exploring the submanifold of minima. In Figure \ref{fig:example-2D} we show the behavior of the system as is, and under our improved dynamics presented in Subsection \ref{sec:dissipation}.

\paragraph{The straightest paths on $\M$.}
The Lagrangian dynamics presented in Equation \eqref{eq:acceleration-euler-lagrange} confirm the well-known fact that geodesics - which are the straightest paths on the manifold - are curves that extremize the kinetic energy integral, since the dynamics coincide with the geodesic equations for $\kappa=0$, i.e. when we disregard the potential energy. A particle with dynamics governed only by kinetic energy is a \emph{free particle}, meaning that it has zero tangential acceleration and constant velocity along the path. Moreover in the case of $\kappa>0$, the resulting trajectories can be interpreted as geodesics on a Finsler manifold under a modified metric:
\begin{equation}
F(\balpha, \dot{\balpha}) = \sqrt{\dot{\balpha}^\top \G{\balpha} \dot{\balpha} + 2 \kappa\cdot\Loss{\balpha}}.
\end{equation}
For further details we refer to textbooks on Riemannian mechanics, e.g. \citet{bucataru2007finsler}.

\paragraph{Particles that never rest.}
A free particle ($\kappa=0$) has constant kinetic energy and no potential energy, and can therefore travel forever on a complete manifold - its trajectory length is unbounded and scales with integration time. Introducing a potential energy term can be interpreted as a gravitational pull on the particle. While this affects the individual energy components along the trajectory, the total energy of the system is conserved; motion up- or downhill on the loss surface simply trades kinetic for potential energy and vice versa. Hence a particle with non-zero initial velocity and dynamics governed by Equation \eqref{eq:acceleration-euler-lagrange} never comes to rest and will instead oscillate around the minimum level set. This is problematic, especially in high-dimensional settings, where the dynamics are both computationally expensive to solve and difficult to simulate over long time horizons. See Appendix \ref{appendix:geodesics} for details.

\subsection{The secret ingredient: energy dissipation through speed-dependent friction}
\label{sec:dissipation}

An obvious way to remove energy from the system - or formally to \emph{dissipate} energy - is through the introduction of external forces, e.g. \citep{minguzzi2015rayleigh, cline2017variational}. In the following we break energy conservation by introducing \emph{friction}; a dissipative force that continuously removes energy from the system. For a dissipation function $\eta(t)>0$, the alternative dynamical system is:
\begin{equation}
    \label{eq:dynamics-with-dissipation}
    \ddot{\balpha} = - \left(\dot{\balpha}^\top \mathbf{H}_{\mathcal{L}}(\balpha)\dot{\balpha} + \kappa\right) \cdot \operatorname{grad}\Loss{\balpha} - \eta(t) \cdot \dot{\balpha},
\end{equation}
for which we describe key properties of the total energy of the system in Lemma \ref{lemma:energy-dissipation} and Theorem \ref{theorem:lasalle-convergence}. We provide proofs for all theorems and lemmas in Appendix \ref{appendix:proof-of-new-theorems}.

\begin{lemma}[Energy dissipation and boundedness]
    \label{lemma:energy-dissipation}
    Let $H(t)=T(\balpha, \dot{\balpha},t) + V(\balpha, t)$ be the the total energy of a particle moving along $\balpha(t)$ at any time $t>t_0$ where $T, V \geq 0$ and assume that $H(t_0) < \infty$. Under the dissipative dynamics in \eqref{eq:dynamics-with-dissipation} where $\eta(t) >0 $ is the dissipation function and holds for all $t \geq t_0$, the total energy dissipates over time as 
    $
        \dot{H} = -2\eta(t)T(\balpha, \dot{\balpha}, t)
    $
    and converges to $H_\infty:=\lim_{t\rightarrow \infty} H(t)$. For all $t\geq t_0$ the total energy is bounded below as: 
    $$
        0 \leq H_\infty \leq H(t) \leq H(t_0) < \infty
    $$
\end{lemma}

\begin{theorem}[Convergence of motion]
    \label{theorem:lasalle-convergence}
    Under the assumptions in Lemma \ref{lemma:energy-dissipation}, and further assuming that the sublevel set $\Omega = \{(\balpha, \dot{\balpha}): H(\balpha, \dot{\balpha})\leq H(t_0)\}$ is compact, all trajectories converge by LaSalle's invariance principle to the largest invariant set
    \[
        \mathcal{I} = \left\{(\balpha, \dot{\balpha}): \dot{\balpha} = \boldsymbol{0}, \ \operatorname{grad}\Loss{\balpha}=\boldsymbol{0}\right\},
    \]
    consisting of equilibria of the loss. Assuming that $\mathcal{L}$ satisfies the strict saddle property, the attracting elements of $\mathcal{I}$ are almost surely local minima and converge to any $\btheta \in S_{\ast}(\mathcal{D})$.
\end{theorem}

\paragraph{A speed-dependent dissipation function.} We suggest defining the dissipation function such that the friction force scales with the speed of the particle, i.e. the square root of the kinetic energy, thereby dissipating energy more at higher speed:
\begin{equation}    
    \label{eq:dissipation-function}
    \eta(t) = \eta_0\sqrt{T(\balpha, \dot{\balpha}, t)} = \eta_0 \lVert\dot{\balpha}(t)\rVert \sqrt{1 + \lVert \grad{\btheta}{\Loss{\balpha(t)}}\rVert^2\cos^2(\omega)}
\end{equation}
where $\eta_0$ is a hyperparameter controlling the effect of the dissipative force and $\omega$ denotes the angle between the velocity $\dot{\balpha}(t)$ and the gradient $\grad{\btheta}{\Loss{\balpha(t)}}$. Together with Equation \eqref{eq:dynamics-with-dissipation}, this dissipation function is what constitutes the sampling dynamics used by \DIMS.

\begin{wrapfigure}{r}{0.425\textwidth}
    \centering
    \vspace{-14pt}
    \includegraphics[width=1.0\linewidth]{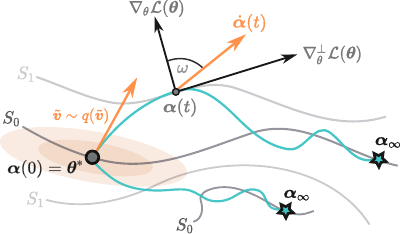}
    \caption{The dynamics of the proposed improved dynamical system. Depending on the initial velocity sample $\tbv\sim q(\tbv)$, \textsc{DiMS} is capable of sampling distinct minimum level sets by dissipating energy particularly when moving proportional to the gradient, and eventually stop. In contrast the geodesic path is unconstrained and requires defining a stop time.}
    \label{fig:dynamics-breakdown}
    \vspace{-25pt}
\end{wrapfigure}

We remark that the dissipation strength directly depends on the Euclidean norm of the intrinsic velocity, $\lVert \dot{\balpha}(t)\rVert$, revealing that the particle slows down faster at high speeds. Furthermore, the dissipation strength depends explicitly on the angle $\omega$ between the velocity of the particle and the steepness of the surface expressed by the gradient. As such, the energy dissipation is maximized when the velocity is proportional to the gradient, i.e. $\dot{\balpha}(t) \propto \grad{\btheta}{\Loss{\balpha(t)}}$, while dissipation only depends on the speed of the Euclidean parameter space curve when the velocity is perpendicular to the gradient, i.e. $\dot{\balpha}(t) \perp \grad{\btheta}{\Loss{\balpha(t)}}$. In other words, the particle is slowed down the least when moving in the direction $\nabla_{\theta}^{\perp}\Loss{\balpha(t)}$ that traverses the current level set.

In Theorems \ref{theorem:convergence-time} and \ref{theorem:gamma-lower-bound} we extend the previous convergence analysis to this specific dissipation function, from which we get a maximal convergence time and an exploration guarantee before convergence.

\begin{theorem}[Convergence time]
    \label{theorem:convergence-time}
    For a particle that traces the curve $\balpha: [t_0, t_1]\rightarrow \Theta$ which converges to $\balpha_{\infty} \in \mathcal{I}$ as $t_1\rightarrow\infty$ according to Theorem \ref{theorem:lasalle-convergence}, let $H(t)$ be the total energy at time $t$ described in Lemma \ref{lemma:energy-dissipation}. Let furthermore $t_{\epsilon} = \inf\{t \geq t_1: T(\balpha, \dot{\balpha}, t) < \epsilon\}$
    be the proxy convergence time at which the kinetic energy is sufficiently low to consider the particle not moving much. Then for dissipation function $\eta(t)=\eta_0 \sqrt{T(\balpha, \dot{\balpha},t)} \geq 0$, an upper bound on the proxy convergence time is
    \[
        t_{\epsilon} \leq t_0 + H(t_0) \left(2\eta_0 \epsilon^{3/2}\right)^{-1}.
    \] 
\end{theorem}

\begin{theorem}[Lower bound on the $\bgamma$-arc-length]
    \label{theorem:gamma-lower-bound}
    Under the assumptions in Theorem \ref{theorem:convergence-time}, and further defining $\Delta \mathcal{L} = \Loss{\balpha\left(t_0\right)} - \Loss{\balpha_{\infty}}$, the arc-length of the curve $\bgamma(t)=h(\balpha) \in\M$ satisfies
    $$
        \frac{T(\balpha, \dot{\balpha}, t_0) + \kappa\Delta \mathcal{L}}{\sqrt{2}\eta_0 H(t_0)} \leq \textsc{Length}(\bgamma).
    $$

\end{theorem}
From Theorem \ref{theorem:convergence-time} we conclude that the time, $t_{\epsilon}$, at which the particle converges to a low energy-state is finite and depends on the initial total energy, friction coefficient and the allowed speed-tolerance $\epsilon$. While the bound may not be tight, it establishes that the dynamics converge in finite time, which is of practical and theoretical significance. Theorem \ref{theorem:gamma-lower-bound} reveals that the particle travels at least a certain distance and that this distance is also based on the initial energy of the system. Hence, sampling with our proposed dynamics guarantees some exploration of the manifold before convergence.

\subsection{How do we obtain samples from $S_{\ast}(\mathcal{D})$?}
Essentially, to generate samples with \DIMS, it suffices to appropriately initialize the dynamical system. Since we want to exploit the modified dynamical system defined in Equations \eqref{eq:dynamics-with-dissipation}-\eqref{eq:dissipation-function} to sample the submanifold of minima $S_{\ast}(\mathcal{D})$, we typically set the particle's initial position $\tilde{\bmu}$ to a minimum solution $\btheta^\ast \in S_{\ast}(\mathcal{D})$ obtained by training a neural network. This choice is purely practical as it reduces run time, and we remark that convergence to $S_{\ast}(\mathcal{D})$ is guaranteed for any initial position.

A point estimate obtained via optimization corresponds to a stable equilibrium of the loss function, meaning that it can not move without an external force acting on it. We therefore need an initial velocity $\tbv$ along which motion is initialized. The particle path $\balpha_{\tilde{\bmu}, \tbv, \eta_0}:[0, t_1]\rightarrow \Theta$ is then computed by numerically integrating the second-order ODE with initial conditions $\balpha(0) = \tilde{\bmu}$ and $\dot{\balpha}(0)=\tbv$. We treat the initial velocity as a random variable and place a distribution over it, $\tbv \sim q(\tbv)$, to ensure variation over the constructed paths. The distribution can freely be chosen, however we define it heuristically as the Laplace approximation; a Gaussian respecting the local loss geometry at $\btheta^\ast$, which is common in approximate Bayesian inference \citep{mackay1992practical, daxberger2021laplace, immer2021improving, bergamin2023riemannian, yu2024riemannian}. We discuss this further in Appendix \ref{appendix:geometric-loop}.

Our sampler generates an implicit distribution over the submanifold of minima and can be formalized: 
\begin{equation}
    \tbv \sim q(\tbv) = \mathcal{N}\left(\boldsymbol{0}, \mathbf{H}_\mathcal{L}(\tilde{\bmu})^{-1}\right) \quad \btheta^{(s)} = \textsc{DiMS}(\tbv | t_1, \eta_0, \tilde{\bmu}, \mathcal{D}) := \balpha_{\tilde{\bmu}, \tbv, \eta_0}(t_1) \in S_{\ast}(\mathcal{D})
\end{equation}
where convergence to $S_{\ast}(\mathcal{D})$ is guaranteed for large enough $t_1$. As pointed out in Theorem \ref{theorem:convergence-time} this time depends on the hyperparameter $\eta_0$ that controls exploration. Flipping the relation, we argue that selecting $\eta_0$ only depends on the computational budget available and that smaller $\eta_0$ are preferred. We describe how to obtain samples with \DIMS computationally in Appendix \ref{appendix:experimental-details}.

\section{Related work}

\paragraph{Loss landscape geometry.}
Understanding the geometry of the loss landscape has been a fundamental research topic in modern machine learning. \citet{pmlr-v70-dinh17b} showed that neural network parameter spaces contain reparameterization invariant sets over which loss properties such as flatness are non-constant, motivating geometry-aware methods that are invariant to such reparameterizations. More recently \citet{modeconnectivity2018garipov} and \citet{pmlr-v80-draxler18a} revealed that independently trained neural networks are connected by low-loss paths, suggesting that the reparameterization invariant set that minimizes the loss is well-described as a connected component. \citet{entezari2021role} conjectured that this connectivity is explained by permutation symmetries within the parameter space, with \citet{pmlr-v267-zhao25i} extending this line of work to continuous symmetries.

\paragraph{Flexible posterior approximations.}
A conceptually simple approximate posterior technique is the Laplace approximation \citep{mackay1992practical} that defines the approximate distribution as a Gaussian with covariance given by the inverse Hessian. In neural network parameter spaces this assumption is often impractical and follow up works has instead used the the Generalized Gauss-Newton as the covariance \citep{daxberger2021laplace} or linearized the function space samples \citep{immer2021improving}. Trajectory-based methods such as SWAG \citep{maddox2019simple} and deep ensembles provide scalable uncertainty estimates but remain agnostic to the geometric structure of the loss landscape. A more principled line of work suggests exploiting the local geometry when defining the approximate posterior; \citet{bergamin2023riemannian} introduced the Riemannian Laplace approximation that defines a distribution in low-loss regions using the pullback metric defined previously, whereas \citet{yu2024riemannian} corrected an inherent bias of the method by replacing the metric with the Fisher metric. In contrast our approach adjusts the bias by modifying the dynamical system. Other works \citep{roy2024reparameterization, fadel2025viking, reichlin2025walking} followed alternative approaches that used a  diffusion-like exploration of the reparameterization invariant solutions in the sampling process. While geometrically principled, these approaches are inherently local and confined to a single loss basin. 

\paragraph{Mechanical systems in optimization and sampling theory.}
First-order optimization methods with momentum such as Polyak's heavy-ball \citep{polyak1964some} and Nesterov's accelerated gradient \citep{nesterov1983method} correspond to discretizations of second-order dissipative ODEs, a connection clarified by \citet{su2016differential} and \citet{kovachki2021continuous}. In the Euclidean setting with constant friction, our proposed continuous-time dynamics recover Polyak's heavy-ball exactly and can thus be seen as a Riemannian generalization of this classical system. Sampling algorithms based on mechanics - such as stochastic gradient Langevin dynamics (SGLD) \citep{welling2011bayesian}, Hamiltonian Monte Carlo (HMC) \citep{neal2011mcmc} or its Lagrangian reformulation \citep{lan2012lagrangian}, along with their Riemannian variants \citep{girolami2011riemann, chen2014stochastic} - are guaranteed to recover the full posterior in the limit of infinite sampling time, which is rarely achieved in practice.

\section{Experiments}

\begin{figure}[tb]
    \centering
    \begin{subfigure}[t]{0.45\textwidth}
        \raggedright
        \includegraphics[width=1\linewidth]{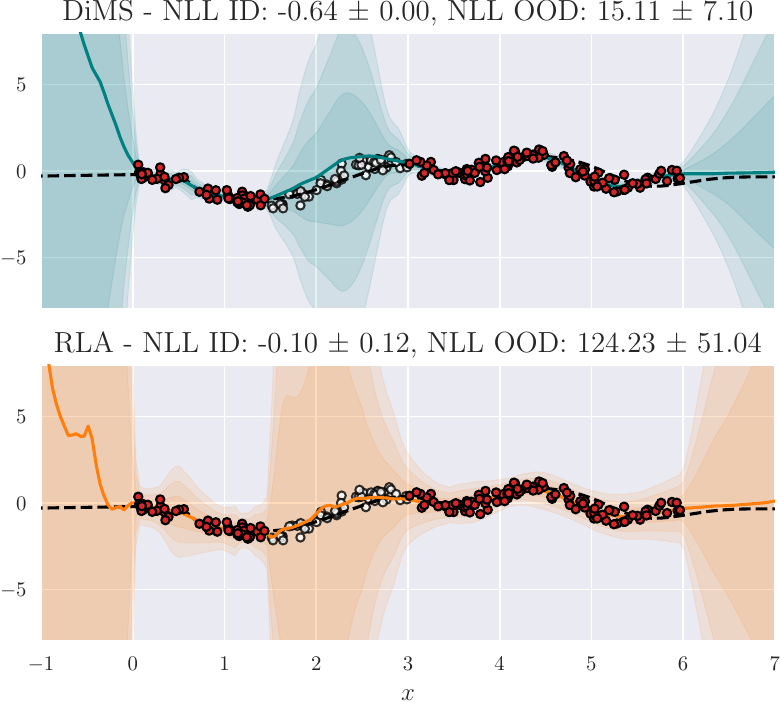}
    \caption{Trained initial $\tilde{\bmu}=\btheta^\ast\in S_{\ast}(\mathcal{D})$}
    \label{fig:snelson-minimum}
    \end{subfigure}%
    \hfill
    \begin{subfigure}[t]{0.45\textwidth}
        \includegraphics[width=1\linewidth]{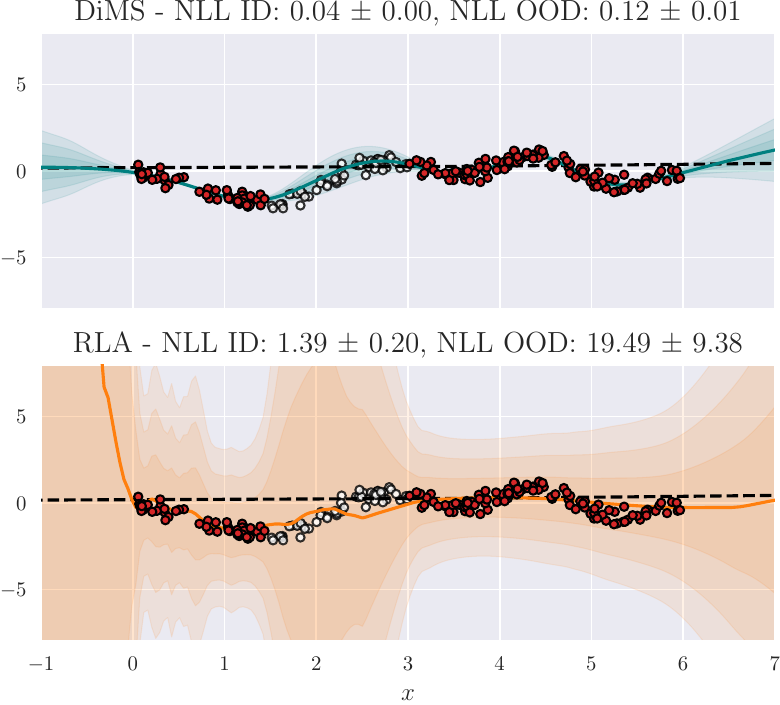}
    \caption{Random initial position $\tilde{\bmu}\not \in S_{\ast}(\mathcal{D})$}
    \label{fig:snelson-non-minimum}
    \end{subfigure}
    \hspace{5pt}
    \caption{Function space samples obtained by \textsc{DiMS} give high OOD estimates without breaking the fit on training data, which \textsc{RLA} does not guarantee. Even when initialized from a suboptimal position, \DIMS provides well-behaved function space samples. Red and white dots are ID and OOD data, respectively, and the black line is the function induced by the initial position parameters. Uncertainty ranges are based on standard deviation and we plot the average over the ensemble as colored lines.
    }
\end{figure}

We evaluate \textsc{DiMS} on uncertainty quantification tasks and compare with methods based on the Laplace approximation, including the sampled (\textsc{LA}), linearized (\textsc{LinLA}) and Riemannian (\textsc{RLA}) Laplace approximations. For all experiments we apply a Bayesian treatment of \emph{all} network parameters and tune the prior precision and likelihood parameters (if any) using the marginal log-likelihood as proposed by \citet{daxberger2021laplace}. We first consider illustrative settings and later extend the analysis to larger neural networks. All experimental details are provided in Appendix \ref{appendix:experimental-details}·

\paragraph{Snelson regression dataset.} We train a fully connected neural network with $2$ hidden layers of $16$ hidden units and \texttt{GELU} activations. We use $20$\% dropout during training to increase the number of active neurons. We set the friction hyperparameter sufficiently low $\eta_0=0.1$ and run \DIMS samples until convergence, whereas \textsc{RLA} per definition stops at $t_1=1$. The resulting predictive uncertainty based on $S=100$ Monte Carlo (MC) samples is shown in Figure \ref{fig:snelson-minimum}. Though \textsc{RLA} is highly uncertain out-of-domain (OOD) it comes at the cost of fitting the data worse in-domain (ID).

As argued by \citet{bergamin2023riemannian} the sampled LA underperforms in regression settings, even when the Hessian used as the covariance matrix is not ill-conditioned and requires either linearization or geometric awareness to obtain a proper fit. However, this depends on the assumption that the starting position is initialized at an optimum. \DIMS is position-agnostic in the sense that the gravitational pull forces the particle toward $S_{\ast}(\mathcal{D})$ no matter the starting position. This setting is similar to a continuous-time gradient descent with momentum and non-zero initial velocity, determining the direction to initialize the optimization process. In Figure \ref{fig:snelson-non-minimum} we show the implication of starting the sampling process from a random initialization of the network with $\eta_0=1.0$ and running until reaching a minimum. We highlight that samples from \DIMS still converge to solutions that fit the training data well and has variation OOD, while \textsc{RLA} samples perform worse under this initialization. Remark also that the OOD variation for \DIMS is more limited when initialized randomly, suggesting that our proposed dynamics allow for maximum exploration when the motion is dominated by the kinetic term rather than the gradient flow dynamics.

\begin{figure}[tb]
    \centering
    \includegraphics[width=1\linewidth]{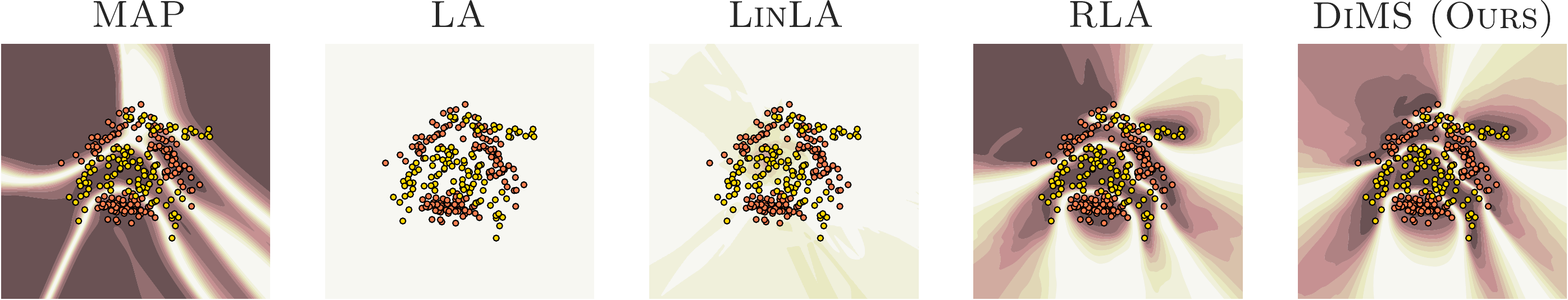}
    \caption{2D binary classification on the banana dataset. The trained \textsc{MAP} model fits the training data well, yet gives unreliable uncertainty estimates. \textsc{LA} and \textsc{LinLA} induce predictive distributions over the input space, yet their approximations are too crude to fit the data properly. In contrast the geometry-aware methods fit the data well while providing better uncertainty estimates than the \textsc{MAP} model, especially \DIMS that captures OOD uncertainty well, better than \textsc{RLA}. Uncertainty is given by the variance of the predictive distribution and we provide calibration metrics in Appendix \ref{appendix:banana-classification}.
    }
    \label{fig:banana-2d}
\end{figure}

\paragraph{Banana classification dataset.} 
To solve the banana classification task depicted in Figure \ref{fig:banana-2d}, we train a fully connected with $4$ hidden layers of $16$ units each, using \texttt{SiLU} activations, $50\%$ dropout and label smoothing of $0.05$, to allow for high OOD variation. We observe that \DIMS exhibits high OOD uncertainty while remaining confident in regions supported by training data and preserving fine-grained details of the data distribution. While \textsc{RLA} performs similarly, it is less uncertain in a large part of the input space. The remaining methods either underfit the data or provide unreliable OOD uncertainty. 

\begin{table}[tb!]
  \centering
  \caption{Negative log-likelihood (NLL $\downarrow$) on UCI classification test sets. We estimate the predictive distribution with $S=30$ samples and report the mean over 5 seeds, including the standard error.}
  \label{tab:uci_nll_test}
  \resizebox{\linewidth}{!}{%
  \begin{tabular}{lcccccc}
    \toprule
     & \textsc{Vehicle} & \textsc{Glass} & \textsc{Ionosphere} & \textsc{Waveform} & \textsc{Australian} & \textsc{Breast C.} \\
     \midrule
    \textsc{LA} & $17.852 \pm 0.175$ & $20.855 \pm 0.299$ & $11.857 \pm 0.691$ & $4.518 \pm 0.218$ & $6.577 \pm 0.231$ & $10.084 \pm 0.479$ \\
    \textsc{LinLA} & $5.446 \pm 0.327$ & $4.097 \pm 0.233$ & $2.591 \pm 0.179$ & $1.287 \pm 0.049$ & $0.789 \pm 0.020$ & $1.662 \pm 0.079$ \\
    \textsc{RLA} & $0.776 \pm 0.051$ & $2.309 \pm 0.192$ & \cellcolor{gray!15}$\mathbf{0.634 \pm 0.066}$ & $0.416 \pm 0.015$ & \cellcolor{gray!15}$\mathbf{0.712 \pm 0.093}$ & $0.185 \pm 0.016$ \\
    \textsc{DiMS (Ours)} & \cellcolor{gray!15}$\mathbf{0.422 \pm 0.015}$ & \cellcolor{gray!15}$\mathbf{1.317 \pm 0.190}$ & $0.724 \pm 0.092$ & \cellcolor{gray!15}$\mathbf{0.299 \pm 0.007}$ & $1.258 \pm 0.109$ & \cellcolor{gray!15}$\mathbf{0.117 \pm 0.013}$ \\
    \bottomrule
  \end{tabular}%
  }
\end{table}

\paragraph{UCI classification datasets.}

We consider $6$ classification problems from the UCI repository and train a classifier for each task, consisting of a $3$-layer network with $16$ hidden units per layer, \texttt{SiLU} activations and $50\%$ dropout. We evaluate these using common uncertainty quantification metrics, specifically the negative log-likelihood (NLL), Brier score, expected calibration error (ECE), and maximum calibration error (MCE). For each seed, metrics are computed by averaging over $S=30$ sampled models, and we report the mean NLL and its standard error across $5$ seeds in Table~\ref{tab:uci_nll_test}. The remaining metrics are provided in Appendix \ref{appendix:uci-classification} and yield similar conclusions.

Overall, \DIMS achieves competitive or superior performance across the datasets, while \textsc{RLA} outperforms \DIMS on $2$ out of the $6$ datasets. Notably, \DIMS was run with a fixed default friction coefficient of $\eta_0=0.5$ across all datasets with no per-dataset tuning -- adjusting this hyperparameter yields better performance than \textsc{RLA} on one of the two datasets where \textsc{RLA} otherwise has an advantage. This suggests that \DIMS is robust to the choice of hyperparameters and can achieve stronger performance with minor tuning effort, as further detailed in Appendix \ref{appendix:uci-classification}.

\paragraph{MNIST and related datasets.} 
We train a convolutional neural network (the LeNet architecture \citep{lecun1989backpropagation}) with approximately $K=44{,}000$ parameters on MNIST and Fashion MNIST and evaluate each method on the test sets along with the test sets of Extended MNIST and KMNIST. We provide the ID and OOD performance in Tables \ref{tab:id-main} and \ref{tab:ood-main}. To define the Laplace approximations we compute the Hessian based on a randomly selected subset containing $N=1{,}000$ of the training samples. We draw $S_B=20$ parameter samples from each method and repeat the sampling process over $B=5$ batches, resulting in a total of $S=100$ samples per method using a friction coefficient of $\eta_0=0.1$. Computing the full Hessian is infeasible for a neural network with a high-dimensional parameter space as it requires inverting a $K\times K$ matrix. We use a low-rank approximation of the Hessian and discuss its implications further in Appendix \ref{appendix:mnist-classification}.

In general, \DIMS is better calibrated and fits ID data better than \textsc{RLA} while sampled \textsc{LA} lacks expressiveness. As expected, \DIMS consistently outperforms the other methods on OOD data.

\begin{table}[tb!]
  \centering
  \caption{In-distribution performance on MNIST and FMNIST with low-rank Hessian approximation.}
  \label{tab:id-main}
  \resizebox{\linewidth}{!}{%
  \begin{tabular}{clccccc}
    \toprule
     &  & \textsc{Accuracy} $\uparrow$ & \textsc{NLL} $\downarrow$ & \textsc{Brier} $\downarrow$ & \textsc{ECE} $\downarrow$ & \textsc{MCE} $\downarrow$ \\
    \cmidrule(lr){2-7}
    \multirow{4}{*}{\rotatebox{90}{\textsc{MNIST}}} & \textsc{LA} & $0.564 \pm 0.009$ & $11.502 \pm 0.507$ & $0.083 \pm 0.002$ & $0.408 \pm 0.009$ & $0.614 \pm 0.006$ \\
     & \textsc{LinLA} & $0.860 \pm 0.007$ & $2.608 \pm 0.151$ & $0.026 \pm 0.001$ & $0.127 \pm 0.006$ & $0.508 \pm 0.007$ \\
     & \textsc{RLA} & \cellcolor{gray!15} $\mathbf{0.953 \pm 0.001}$ & $0.609 \pm 0.009$ & $0.009 \pm 0.0001$ & $0.042 \pm 0.0005$ & $0.472 \pm 0.008$ \\
     & \textsc{DiMS (ours)} & $0.950 \pm 0.0002$ & \cellcolor{gray!15}$\mathbf{0.271 \pm 0.002}$ & \cellcolor{gray!15}$\mathbf{0.008 \pm 4\cdot10^{-5}}$ & \cellcolor{gray!15}$\mathbf{0.035 \pm 0.0002}$ & \cellcolor{gray!15}$\mathbf{0.379 \pm 0.009}$ \\
    \cmidrule(lr){2-7}
    \multirow{4}{*}{\rotatebox{90}{\textsc{FMNIST}}} & \textsc{LA} & $0.147 \pm 0.003$ & $53.519 \pm 1.298$ & $0.167 \pm 0.001$ & $0.825 \pm 0.003$ & $0.866 \pm 0.003$ \\
     & \textsc{LinLA} & $0.385 \pm 0.009$ & $11.636 \pm 0.270$ & $0.115 \pm 0.002$ & $0.555 \pm 0.008$ & $0.647 \pm 0.006$ \\
     & \textsc{RLA} & $0.767 \pm 0.002$ & $9.210 \pm 1.266$ & $0.045 \pm 0.0004$ & $0.220 \pm 0.002$ & $0.520 \pm 0.006$ \\
     & \textsc{DiMS (ours)} & \cellcolor{gray!15}$\mathbf{0.814 \pm 0.002}$ & \cellcolor{gray!15}$\mathbf{1.443 \pm 0.010}$ & \cellcolor{gray!15}$\mathbf{0.033 \pm 0.0003}$ & \cellcolor{gray!15}$\mathbf{0.149 \pm 0.001}$ & \cellcolor{gray!15}$\mathbf{0.414 \pm 0.021}$ \\
    \bottomrule
  \end{tabular}%
  }
  \vspace{-9pt}
\end{table}

\begin{table}[tb!]
  \centering
  \caption{Out-of-distribution performance measured as AUROC $\uparrow$.}
  \label{tab:ood-main}
  \resizebox{\linewidth}{!}{%
  \begin{tabular}{lcccccc}
    \toprule
     Trained on: & \multicolumn{3}{c}{\leavevmode\leaders\hrule height .55ex depth -.45ex\hfill\kern0pt\enspace\textsc{MNIST}\enspace\leavevmode\leaders\hrule height .55ex depth -.45ex\hfill\kern0pt} & \multicolumn{3}{c}{\leavevmode\leaders\hrule height .55ex depth -.45ex\hfill\kern0pt\enspace\textsc{FMNIST}\enspace\leavevmode\leaders\hrule height .55ex depth -.45ex\hfill\kern0pt} \\
     Evaluated on: & \textsc{EMNIST} & \textsc{FMNIST} & \textsc{KMNIST} & \textsc{EMNIST} & \textsc{KMNIST} & \textsc{MNIST} \\
    \midrule
    \textsc{LA} & $0.638 \pm 0.006$ & $0.745 \pm 0.006$ & $0.681 \pm 0.007$ & $0.432 \pm 0.005$ & $0.436 \pm 0.005$ & $0.433 \pm 0.005$ \\
    \textsc{LinLA} & $0.728 \pm 0.005$ & $0.829 \pm 0.006$ & $0.775 \pm 0.006$ & $0.427 \pm 0.003$ & $0.427 \pm 0.003$ & $0.430 \pm 0.003$ \\
    \textsc{RLA} & $0.846 \pm 0.004$ & $0.939 \pm 0.003$ & $0.882 \pm 0.004$ & $0.556 \pm 0.004$ & $0.586 \pm 0.004$ & $0.574 \pm 0.004$ \\
    \textsc{DiMS (ours)} & \cellcolor{gray!15}$\mathbf{0.961 \pm 0.0002}$ & \cellcolor{gray!15}$\mathbf{0.971 \pm 0.0003}$ & \cellcolor{gray!15}$\mathbf{0.979 \pm 0.0002}$ & \cellcolor{gray!15}$\mathbf{0.715 \pm 0.002}$ & \cellcolor{gray!15}$\mathbf{0.779 \pm 0.001}$ & \cellcolor{gray!15}$\mathbf{0.781 \pm 0.003}$ \\
    \bottomrule
  \end{tabular}%
  }
\end{table}

\section{Conclusion}

We proposed \DIMS, a sampler grounded in dissipative mechanics on Riemannian manifolds that is guaranteed to explore the minimum-loss submanifolds of neural networks, i.e. parameter space regions where models perform well on training data while retaining high predictive uncertainty on out-of-distribution inputs.
We provided theoretical guarantees that the dynamics converge to these minimum-loss submanifolds in finite time regardless of the initial conditions, and validated the sampler empirically across various settings.
\DIMS consistently improves over or performs on par with competing Laplace-based methods on uncertainty quantification tasks, while being robust to its single physically-motivated hyperparameter --the friction coefficient-- which governs energy dissipation in the dynamical system. We hope this work encourages further research on geometry and physics inspired approaches to minima exploration and uncertainty quantification in deep learning.

\textbf{Limitations and future work.}\quad
\DIMS is, by design, a geometry-based approach and therefore computationally demanding, as is typical for such methods. The main cost stems from solving an ODE in a high-dimensional space, which could potentially be mitigated using tailored or stochastic solvers. While \DIMS is robust to the friction hyperparameter, a position-dependent formulation could be considered. Sampling reparameterization invariant models requires a sufficiently expressive function space, and it is of interest to characterize the conditions under which this holds. 

\begin{ack}
This work was supported by the Danish Data Science Academy, which is funded by the Novo Nordisk Foundation (NNF21SA0069429) and VILLUM FONDEN (40516), and by the DFF Sapere Aude Starting Grant ``GADL''.
\end{ack}

\medskip
\bibliographystyle{plainnat}
{\small
    \bibliography{references}
}


\newpage
\appendix
\newpage
\appendix

\section{Details on the loss manifold}
\renewcommand{\theequation}{A.\arabic{equation}}
\setcounter{equation}{0} 

\subsection{Quantities of the pullback metric required for the derivations}
The metric is given by:
\begin{equation}
    \G{\btheta} = \I_K + \grad{\btheta}{\Loss{\btheta}} \grad{\btheta}{\Loss{\btheta}}^\top,
\end{equation}
The inverse metric and its square root follows from the Sherman-Morisson formula:
\begin{equation}
    \invG{\btheta} = \I_K - \frac{\grad{\btheta}{\Loss{\btheta}} \grad{\btheta}{\Loss{\btheta}}^\top}{1 + \left\lVert \grad{\btheta}{\Loss{\btheta}}\right\rVert^2 },
\end{equation}
\begin{equation}
    \sqrtinvG{\btheta} = \I_K - \frac{\grad{\btheta}{\Loss{\btheta}} \grad{\btheta}{\Loss{\btheta}}^\top}{1 + \left\lVert \grad{\btheta}{\Loss{\btheta}}\right\rVert^2 +\sqrt{1 + \left\lVert \grad{\btheta}{\Loss{\btheta}}\right\rVert^2}}.
\end{equation}
The Riemannian gradient can then be expressed as:
\begin{equation}
    \operatorname{grad}\Loss{\btheta} = \invG{\btheta} \grad{\btheta}{\Loss{\btheta}} = \frac{ \grad{\btheta}{\Loss{\btheta}}}{1 + \lVert \grad{\btheta}{\Loss{\btheta}}\rVert^2}.
\end{equation}
The Christoffel symbols are needed to solve the Euler-Lagrange equations. These describe how coordinate basis vectors change along the manifold and determine how tangent vectors are corrected when moving on the manifold. We refer to classic textbooks on differential geometry, e.g. \cite{do1992riemannian, lee2018introduction}, for further reading. For the specific definition of the manifold, the Christoffel symbols simplify to:
\begin{equation}
    \label{eq:christoffel-monge}
    \Gamma_{ij}^k(\btheta) = \left(\frac{\partial^2}{\partial \theta_i \partial \theta_j} \Loss{\btheta}\cdot \sum_{m}^K \mathbf{G}_{km}^{-1}(\btheta) \frac{\partial}{\partial \theta_m} \Loss{\btheta}\right).
\end{equation}
When computing the Christoffel symbols, we leveraged the closed form expression of the derivative of the metric that is the outer product of the Hessian columns with the gradient of the loss. By letting $\mathbf{H}_{\mathcal{L}}^{(k)}(\btheta)$ denote the $k$'th column of the Hessian of $\Loss{\btheta}$, the derivative of the metric is:
\begin{equation}
    \frac{\partial}{\partial \theta_k} \G{\btheta} = \mathbf{H}_{\mathcal{L}}^{(k)}(\btheta) \grad{\btheta}{\Loss{\btheta}}^\top + \grad{\btheta}{\Loss{\btheta}} \mathbf{H}_{\mathcal{L}}^{(k)}(\btheta)^\top.
\end{equation}
For defining how the energy of a particle evolves over time, we need a description of how the metric evolves along the curve. We therefore establish that the time-derivative of the metric along $\balpha(t)$ is:
{\small
\begin{equation}
    \dot{\mathbf{G}}(\balpha) = \sum_{k=1}^K \dot{\alpha}_k \frac{\partial}{\partial \theta_k} \G{\balpha} = \mathbf{H}_{\mathcal{L}}(\balpha) \dot{\balpha} \grad{\btheta}{\Loss{\balpha}}^\top + \grad{\btheta}{\Loss{\balpha}} \dot{\balpha}^\top \mathbf{H}_{\mathcal{L}}(\balpha)
\end{equation}
}
Furthermore, the $\mathbf{G}$-norm of the gradient is:
\begin{equation}
    \lVert \grad{\btheta}{\Loss{\btheta}}\rVert_{\G{\btheta}} = \sqrt{\grad{\btheta}{\Loss{\btheta}}^\top \G{\btheta} \grad{\btheta}{\Loss{\btheta}}} = \sqrt{\lVert \grad{\btheta}{\Loss{\btheta}} \rVert^2 + \lVert \grad{\btheta}{\Loss{\btheta}}\rVert^4}.
\end{equation}

\subsection{Geometry-driven posterior approximations on the loss manifold}
\label{appendix:geometric-loop}

We consider the manifold $\M$ defined as the graph of the negative log-posterior loss on the training data as in Equation \eqref{eq:log-posterior-loss}. For posterior approximations defined as the pushforward of a Gaussian in the tangent space via some map, we provide a schematic overview of the relations between the ambient tangent space $\TM{\bmu}$, intrinsic tangent space $\mathcal{T}_{\tilde{\bmu}}\Theta$, parameter space $\Theta$ and the manifold $\M$. The schematic is meant as a side-kick to what we depicted in Figure \ref{fig:curve-intuition}. We remark that we do not requires the pushforward map to be a diffeomorphism, yet remark that for non-bijective maps, the inverse mapping from samples to the initial position does not exist. This limits obtaining an explicit approximate posterior via the change-of-variables formula, yet admits an implicit approximate posterior obtained through sampling. 
\begin{center}
    \begin{tikzpicture}[>=stealth, shorten <=15pt, shorten >=15pt]
        
        \coordinate (BL) at (0,0);
        \coordinate (BR) at (4,0);
        \coordinate (TR) at (4,2);
        \coordinate (TL) at (0,2);
        
        \draw (BL) node {$\mathcal{T}_{\tilde{\bmu}}\Theta$};
        \draw (BR) node {$\Theta$};
        \draw (TR) node {$\M$};
        \draw (TL) node {$\TM{\bmu}$};
        
        \draw[->] (BL) to[bend left=0]
            node[midway, above] {$\psi$} (BR);
        \draw[->, dashed] (BR) to[bend left=20]
            node[midway, below] {$\left(\psi^{-1}\right)$} (BL);
                
        \draw[->] (TR) to[bend left=30]
            node[midway, right] {$h^{-1}$} (BR);
        \draw[->] (BR) to[bend left=0]
            node[midway, left] {$h$} (TR);
                
        \draw[<-, dashed] (TL) to[bend left=20]
            node[midway, above] {$\left(\phi^{-1}\right)$} (TR);
        \draw[<-] (TR) to[bend left=0]
            node[midway, below] {$\phi$} (TL);
                
        \draw[<-] (BL) to[bend left=30]  
            node[midway, left] {$\mathbf{J}_{h}^{\dagger}(\bmu)$} (TL);
        \draw[<-] (TL) to[bend left=0]  
            node[midway, right] {$\J{h}{\tilde{\bmu}}$} (BL);
    \end{tikzpicture}
\end{center}

The speed of a curve $\bgamma(t) \in \M$ at time $t$ is given by the norm of its velocity, $\left\lVert \dot{\bgamma}(t)\right \rVert$. Recall that an element of the ambient tangent space can be expressed in intrinsic coordinates as $\J{h}{\btheta}\tbv$, and using the Moore-Penrose inverse, $\J{h}{\btheta}^\dagger = \left(\J{h}{\btheta}^\top \J{h}{\btheta}\right)^{-1} \J{h}{\btheta}^\top = \invG{\btheta} \J{h}{\btheta}^\top$ we can hence map velocity samples from the ambient tangent space to their parameter space coordinates, since the mapping is preserved:
\begin{equation}
    \tbv = \J{h}{\btheta}^\dagger \bv = \J{h}{\btheta}^\dagger \J{h}{\btheta} \tbv = \invG{\btheta}\G{\btheta} \tbv = \tbv.
\end{equation}
Hence, mapping a vector from the parameter space coordinates to its tangent space coordinates preserves the mapping. We then define the distribution of initial velocities directly in the parameter space coordinates as:
\begin{equation}
    \tbv = \sqrtinvG{\btheta} \beps, \qquad q(\beps)=\mathcal{N}\left(\boldsymbol{0}, \mathbf{\Sigma}\right)
\end{equation}
due to linearity of Gaussians. We see that the initial speed samples from this distribution is invariant to the position $\bmu$, since:
\begin{equation}
    \lVert\bv\rVert^2 = \lVert \tbv \rVert^2_{\G{\btheta}} = \beps^\top \underset{=\I_K}{\underbrace{\sqrtinvG{\btheta} \G{\btheta} \sqrtinvG{\btheta}}} \  \beps =  \lVert \beps \rVert^2,
\end{equation}
which is not the case if we disregard precoditioning by the square-root of the inverse metric, see:
\begin{equation}
    \lVert\bv\rVert^2 = \lVert \tbv \rVert^2_{\G{\btheta}} = \tbv^\top \G{\btheta} \tbv.
\end{equation}
Remark that at a local minimum $\btheta^\ast$ the preconditioning has no effect since $\sqrtinvG{\btheta^\ast} = \I_K$ by definition. For geometry-driven approximate posteriors that initialize the dynamics at a minimum - which is the case for any such distribution based on the Laplace approximation - the preconditioning does not need to be accounted for.

\section{Derivations related to curve energy}
\label{appendix:geodesics}
\renewcommand{\theequation}{B.\arabic{equation}}
\setcounter{equation}{0} 
For a unit-mass particle traveling on the loss manifold, its energy is fully captured by a kinetic and a potential energy component. The kinetic energy of a particle moving on the manifold can equally be formulated in its intrinsic coordinates, here referring to the parameter space. These formulations relate as:
\begin{equation}
    \label{eq:kinetic}
    E_{\text{kin}}=T(\bgamma, \dot{\bgamma}) = \frac{1}{2} \cdot g_{\bgamma}(\dot{\bgamma}, \dot{\bgamma}) = \frac{1}{2} \cdot \dot{\balpha}^\top \G{\balpha}\dot{\balpha} = T(\balpha, \dot{\balpha}) \geq 0,
\end{equation}
Similarly, the potential energy of such a system is described as:
\begin{equation}
    \label{eq:potential}
    E_{\text{pot}} = V(\bgamma) =  \kappa\cdot\left(\mathcal{L} \circ h^{-1}\right)(\bgamma) - V^\ast = \kappa \cdot \left(\Loss{\balpha} - \Loss{\btheta^\ast_{\text{glob}}}\right) = V(\balpha) \geq0,
\end{equation}
where $\btheta^\ast_{\text{glob}}$ is a parameter configuration associated to the minimum potential $V^\ast$ and $\kappa>0$ is a positive constant summarizing the gravitational acceleration and the particle's mass. The particle dynamics are then described through the Lagrangian that encodes the energy balance between the kinetic and potential energies:
\begin{equation}
    \label{eq:lagrangian}
    L(\bgamma, \dot{\bgamma}) = T(\bgamma, \dot{\bgamma}) - V(\bgamma) = T(\balpha, \dot{\balpha}) - V(\balpha) = L(\balpha, \dot{\balpha}).
\end{equation}
An extremal action curve is a trajectory of the particle, i.e. $\bgamma: [t_0, t_1] \rightarrow \mathcal{M}$, that respects the principle of stationary action, meaning that it minimizes an action functional $S$ defined by:
\begin{equation}    
    \label{eq:stationary-action}
    \underset{\bgamma}{\arg \min} \ S = \underset{\bgamma}{\arg \min}\int_{t_0}^{t_1} L(\bgamma, \dot{\bgamma}) dt 
    = \underset{\balpha}{\arg \min} \int_{t_0}^{t_1} L(\balpha, \dot{\balpha}) dt.
\end{equation}
What should be evident from Equations \eqref{eq:kinetic}-\eqref{eq:stationary-action} is that energies and dynamics of a particle moving on the loss manifold can equivalently be expressed and computed by considering the parameter space trajectory of the particle.

A curve $\balpha(t)$ is a stationary point of $S$ if and only if it satisfies the Euler-Lagrange equations:
\begin{equation}
    \frac{\partial L}{\partial \alpha_k} - \frac{d}{dt} \frac{\partial L}{\partial \dot{\alpha}_k} = 0.
\end{equation}

\subsection{Solution to the Euler-Lagrange equations}
Inserting the definition of the Lagrangian from Equation \eqref{eq:lagrangian} into the Euler-Lagrange equations, we get:
\begin{equation}    
    \frac{\partial T}{\partial \alpha_k} -  \frac{\partial V}{\partial \alpha_k} - \frac{d}{dt} \left(\frac{\partial T}{\partial \dot{\alpha}_k} - \frac{\partial V}{\partial \dot{\alpha}_k}\right) = 0,
\end{equation}
and since the potential is a function of the coordinates only, this simplifies to:
\begin{equation}
    \frac{\partial T}{\partial \alpha_k}  - \frac{d}{dt} \frac{\partial T}{\partial \dot{\alpha}_k}  = \frac{\partial V}{\partial \alpha_k}.
\end{equation}
Remark the coordinate dependency of the kinetic energy in the first term, a result of the fact that the particle is moving on the manifold with metric $\G{\btheta}$. We insert the kinetic and potential energies and get:
\begin{align}
    \label{eq:previous-equation}
    \frac{1}{2} \left(\frac{\partial}{\partial \alpha_k} \left(\dot{\balpha}^\top \G{\balpha} \dot{\balpha}\right) - \frac{d}{dt} \frac{\partial}{\partial \dot{\alpha}_k} \left(\dot{\balpha}^\top \G{\balpha} \dot{\balpha}\right) \right)  = \kappa \cdot \frac{\partial}{\partial \alpha_k}\Loss{\balpha}.
\end{align}
We proceed by considering the derivative terms individually. First realize that:
\begin{equation}
    \label{eq:first-deriv}
    \frac{\partial}{\partial \alpha_k} \left(\dot{\balpha}^\top \G{\balpha} \dot{\balpha}\right) = \frac{\partial \mathbf{G}_{ij}(\balpha)}{\partial \alpha_k} \dot{\alpha}_i \dot{\alpha}_j.
\end{equation}
Next, we leverage the product and the chain rule to get:
\begin{align}
    \label{eq:second-deriv}
    \frac{d}{dt} \frac{\partial}{\partial \dot{\alpha}_k} \left(\dot{\balpha}^\top \G{\balpha} \dot{\balpha}\right) &= 2 \frac{d}{dt} \mathbf{G}_{kj}(\balpha) \dot{\alpha}_j\\ \notag
    &= 2 \left(\frac{d}{dt} \mathbf{G}_{kj}(\balpha)\right) \dot{\alpha}_j + 2  \mathbf{G}_{kj}(\balpha) \ddot{\alpha}_j \\ \notag
    &= 2 \frac{\partial \mathbf{G}_{kj}(\balpha)}{\partial\alpha_i}\dot{\alpha}_i \dot{\alpha}_j + 2  \mathbf{G}_{kj}(\balpha) \ddot{\alpha}_j \\ \notag
    &= \left(\frac{\partial \mathbf{G}_{kj}(\balpha)}{\partial\alpha_i} + \frac{\partial \mathbf{G}_{ki}(\balpha)}{\partial\alpha_j} \right) \dot{\alpha}_i \dot{\alpha}_j + 2  \mathbf{G}_{kj}(\balpha) \ddot{\alpha}_j, 
\end{align}
and inserting Equations \eqref{eq:first-deriv}-\eqref{eq:second-deriv} into Equation \eqref{eq:previous-equation} gives:
{\small
\begin{equation}
    \label{eq:b11}
    \frac{1}{2} \left(\frac{\partial \mathbf{G}_{ij}(\balpha))}{\partial \alpha_k} - \frac{\partial \mathbf{G}_{kj}(\balpha))}{\partial\alpha_i} - \frac{\partial \mathbf{G}_{ki}(\balpha))}{\partial\alpha_j}\right)\dot{\alpha}_i \dot{\alpha}_j - \mathbf{G}_{kj}(\balpha) \ddot{\alpha}_j = \kappa \cdot\frac{\partial}{\partial \alpha_k}\Loss{\balpha}.
\end{equation}
}
To find the equations of motion, we isolate the acceleration. To do so, realize that: 
\begin{equation}
    \sum_k \mathbf{G}^{-1}_{lk}(\balpha) \mathbf{G}_{kj}(\balpha) = \delta_j^l,
\end{equation}
which allows for rewriting Equation \eqref{eq:b11} using the Christoffel symbols:
\small{
\begin{align}
    \frac{1}{2}  \mathbf{G}^{-1}_{lk}(\balpha) \left(\frac{\partial \mathbf{G}_{kj}(\balpha)}{\partial\alpha_i} + \frac{\partial \mathbf{G}_{ki}(\balpha)}{\partial\alpha_j} - \frac{\partial \mathbf{G}_{ij}(\balpha)}{\partial \alpha_k}\right)\dot{\alpha}_i \dot{\alpha}_j + \ddot{\alpha}_l &= -\kappa \cdot \mathbf{G}^{-1}_{lk}(\balpha) \cdot \frac{\partial}{\partial \alpha_k}\Loss{\balpha}\\
    \Rightarrow \qquad \qquad \Gamma_{ij}^l(\balpha) \dot{\alpha}_i \dot{\alpha}_j + \ddot{\alpha}_l &= -\kappa \cdot \mathbf{G}^{-1}_{lk}(\balpha) \cdot \frac{\partial}{\partial \alpha_k}\Loss{\balpha},
\end{align}
}
from which the acceleration can be isolated and forms a second-order ordinary differential equation (ODE):
\begin{equation}
    \ddot{\alpha}_l = -\Gamma_{ij}^l(\balpha) \dot{\alpha}_i \dot{\alpha}_j - \kappa \cdot \mathbf{G}^{-1}_{lk}(\balpha) \cdot \frac{\partial}{\partial \alpha_k}\Loss{\balpha},
\end{equation}
By swapping indices and recalling that the term coming from the potential energy equals the Riemannian gradient, the expression becomes:
\begin{equation}
    \ddot{\alpha}_k = - \Gamma_{ij}^k(\balpha) \dot{\alpha}_i \dot{\alpha}_j - \kappa \cdot \operatorname{grad}\Loss{\balpha}_k,
\end{equation}
where $\operatorname{grad}\Loss{\balpha}_k$ denotes the $k$-th index of the Riemannian gradient. Remark that we used no assumptions about the metric and that these dynamics hold for a particle traveling on any Riemannian manifold.

\paragraph{Solution under the pullback metric.}
By leveraging the Christoffel symbols for the loss manifold provided in Equation \eqref{eq:christoffel-monge}, it is straight-forward to show that the Euler-Lagrange equation from above simplifies to:
\begin{equation}
    \label{eq:b17}
    \ddot{\balpha} = - \frac{\dot{\balpha}^\top \mathbf{H}_{\mathcal{L}}(\balpha)\dot{\balpha}}{1 + \left \lVert \grad{\btheta}{\Loss{\balpha}}\right \rVert^2} \cdot \grad{\btheta}{\Loss{\balpha}} - \kappa \cdot \operatorname{grad}\Loss{\balpha} = - \left(\dot{\balpha}^\top \mathbf{H}_{\mathcal{L}}(\balpha)\dot{\balpha}+\kappa\right)\cdot\operatorname{grad}\Loss{\balpha}.
\end{equation}

\paragraph{Interpreting the potential term as a gravitational pull.}
The dynamics derived in the previous section also appear when adding a gravitational force component to the tangential acceleration of the particle traveling on $\M$ along $\bgamma(t)$. Leveraging the covariant derivative, see that:
\begin{equation}
    \nabla_{\dot{\bgamma}} \dot{\bgamma} = \boldsymbol{d} - \langle\boldsymbol{d},\hat{\boldsymbol{n}}(\balpha)\rangle \hat{\boldsymbol{n}}(\balpha) = \begin{bmatrix}
        -\kappa \cdot \operatorname{grad}\Loss{\balpha}, & \kappa \cdot \frac{\lVert \grad{\btheta}{\Loss{\balpha}}\rVert^2}{1+ \lVert \grad{\btheta}{\Loss{\balpha}}\rVert^2}
    \end{bmatrix}^\top,
\end{equation}
where $\boldsymbol{d}=\left[0, \dots, 0, -\kappa\right]^\top \in \RR^{K+1}$ is the gravitational pull and the normal vector to the loss surface is given by $\hat{\boldsymbol{n}}(\balpha)=\left[-\grad{\btheta}{\Loss{\balpha}}, 1 \right]^\top \in \RR^{K+1}$. Considering all parameter space dimensions, we see that this corresponds to the term originating from the potential energy when solving the Euler-Lagrange equations.

\paragraph{Gravity-influenced curves are geodesics on Finsler manifolds.}
As argued by \citet{bucataru2007finsler}, a Finslerian mechanical system with metric $F(\balpha, \dot{\balpha}, t)$ corresponds to defining the Lagrangian as:
\begin{equation}
    L(\balpha, \dot{\alpha}, t) = \frac{1}{2}F(\balpha, \dot{\balpha},t)^2.
\end{equation}
Inverting the relation and using the definition of the Lagrangian as the energy balance between the kinetic and potential energy, it is easy to see that the proposed dynamics from Equation \eqref{eq:b17} correspond to motion on a Finsler manifold with metric:
\begin{equation}
    F(\balpha, \dot{\balpha},t) = \sqrt{\dot{\balpha}(t)^\top \G{\balpha(t)} \dot{\balpha}(t) + 2 \kappa\cdot\Loss{\balpha(t)}}.
\end{equation}

\subsection{Energy flow equations under the pullback metric}
\label{appendix:energy-flow}

We consider a parameter space curve $\balpha(t)$ induced by the non-dissipative dynamics stated in Equation \eqref{eq:b17}. In this section we will show the well-known fact that a non-dissipative system conserves energy by trading of kinetic with potential energy and vice versa and later show the implication of introducing dissipation as a friction force. First, we take the time-derivative of the kinetic energy:
\begin{align}
    \frac{d}{dt} T(\balpha, \dot{\balpha}) = \frac{1}{2} \frac{d}{dt} \left(\dot{\balpha}^\top \G{\balpha} \dot{\balpha}\right) = \dot{\balpha}^\top \G{\balpha} \ddot{\balpha} + \frac{1}{2} \dot{\balpha}^\top \dot{\mathbf{G}}(\balpha) \dot{\balpha}.
\end{align}
We study the two terms separately. For the first term, we leverage the definition of the acceleration from Equation \eqref{eq:b17} to get:
\begin{align}
    \dot{\balpha}^\top \G{\balpha} \ddot{\balpha} &= - \frac{\dot{\balpha}^\top \mathbf{H}_{\mathcal{L}}(\balpha)\dot{\balpha}}{1 + \left \lVert \grad{\btheta}{\Loss{\balpha}}\right \rVert^2} \cdot \dot{\balpha}^\top \G{\balpha}\grad{\btheta}{\Loss{\balpha}} - \kappa\cdot \dot{\balpha}^\top \grad{\btheta}{\Loss{\balpha}} \\ \notag
    &= -\dot{\balpha}^\top \mathbf{H}_{\mathcal{L}}(\balpha)\dot{\balpha} \cdot \dot{\balpha}^\top \G{\balpha}\invG{\balpha} \grad{\btheta}{\Loss{\balpha}} - \kappa\cdot \dot{\balpha}^\top \grad{\btheta}{\Loss{\balpha}} \\ \notag
    &= -\left(\dot{\balpha}^\top \mathbf{H}_{\mathcal{L}}(\balpha)\dot{\balpha} + \kappa\right)\cdot \dot{\balpha}^\top \grad{\btheta}{\Loss{\balpha}}.
\end{align}
For the second term, we insert the definition of the time-derivative of the metric and get:
\begin{align}
    \frac{1}{2} \dot{\balpha}^\top \dot{\mathbf{G}}(\balpha) \dot{\balpha} = \dot{\balpha}^\top \mathbf{H}_{\mathcal{L}}(\balpha) \dot{\balpha}\cdot \dot{\balpha}^\top \grad{\btheta}{\Loss{\balpha}}.
\end{align}
Combining the two terms and applying the chain rule inversely, we obtain the \emph{law of conservation of energy} which is well-known from physics and exactly reveals the direct relation between kinetic and potential energy for the moving particle:
\begin{equation}
    \label{eq:energy-conservation}
    \dot{T}(\balpha, \dot{\balpha}) =\frac{d}{dt} T(\balpha, \dot{\balpha}) = - \kappa \cdot \dot{\balpha}^\top \grad{\btheta}{\Loss{\balpha}} = - \kappa \cdot \frac{d}{dt} \Loss{\balpha} = - \frac{d}{dt} V(\balpha) = - \dot{V}(\balpha).
\end{equation}

\paragraph{Introducing dissipation through friction.}
The introduction of dissipation as a friction force based on dissipation function $\eta(t)>0$ results in a modified version of the particle's acceleration:
\begin{equation}
    \label{eq:dissipative-dynamics-appendix}
    \ddot{\balpha} = - \left(\dot{\balpha}^\top \mathbf{H}_{\mathcal{L}}(\balpha)\dot{\balpha} + \kappa \right)\cdot\operatorname{grad}\Loss{\balpha} - \eta(t) \cdot \dot{\balpha}.
\end{equation}
As a consequence the the time-derivative of the kinetic energy changes. Inserting the dissipative acceleration equation into the previous derivation, it is straight-forward to see that:
\begin{align}
    \label{eq:energy-flow-dissipation-appendix}
    \dot{T}(\balpha, \dot{\balpha}) &= - \dot{V}(\balpha) - \eta(t)\cdot \dot{\balpha}^\top \G{{\balpha}} \dot{\balpha} \\ \notag
    &= - \dot{V}(\balpha) - 2\eta(t)\cdot T(\balpha, \dot{\balpha}).
\end{align}
Since $\eta(t)\cdot \dot{\balpha}^\top \G{{\balpha}} \dot{\balpha} \geq 0$ by definition, the kinetic energy gradually decreases even though a drop in potential energy can still locally increase the kinetic energy. 

Lastly, we rewrite the speed-dependent dissipation function considered in our sampler \DIMS as:
\begin{align}
    \eta(t) = \eta_0\sqrt{T(\balpha, \dot{\balpha})} &= \eta_0\sqrt{\dot{\balpha}^\top \G{\balpha}\dot{\balpha}} \\ \notag
    &=\eta_0 \sqrt{\lVert\dot{\balpha}\rVert^2 + \lVert \dot{\balpha}^\top \grad{\btheta}{\Loss{\balpha}}\rVert^2} \\ \notag
    &= \eta_0\sqrt{\lVert\dot{\balpha}\rVert^2 + \left(\lVert \dot{\balpha}\rVert \lVert \grad{\btheta}{\Loss{\balpha}}\rVert \cos(\omega)\right)^2} \\ \notag
    &= \eta_0\lVert\dot{\balpha}\rVert \sqrt{1 + \lVert \grad{\btheta}{\Loss{\balpha}}\rVert^2 \cos^2(\omega)},
\end{align}
where $\eta_0 > 0$ is the friction coefficient, treated as a hyperparameter. This clearly shows that the dissipation depends on the angle between the velocity and the gradient, $\omega$, for the position-velocity state at time $t$.

\section{Proofs of lemmas and theorems}
\renewcommand{\theequation}{C.\arabic{equation}}
\setcounter{equation}{0} 

\label{appendix:proof-of-theorems}
\label{appendix:proof-of-new-theorems}

\paragraph{Lemma \ref{lemma:energy-dissipation} - energy dissipation and boundedness.} 

Consider a curve $\balpha: [t_0,\infty] \rightarrow \Theta$ following the general dissipative dynamics presented in Equation \eqref{eq:dissipative-dynamics-appendix}, and recall that by definition the kinetic and potential from Equations \eqref{eq:kinetic}-\eqref{eq:potential}, the energy terms are bounded below as $T(\balpha,\dot{\balpha}, t), V(\balpha, t) \geq 0$. We define the total energy of the particle at time $t \geq t_0$ as the Hamiltonian:
\begin{equation}
    H(t):=H(\balpha, \dot{\balpha}, t) = T(\balpha, \dot{\balpha}, t) + V(\balpha, t) \geq 0.
\end{equation}
From Equation \eqref{eq:energy-flow-dissipation-appendix} and given that $\eta(t) > 0$ it holds that:
\begin{align}
    \dot{H}(\balpha, \dot{\balpha},t) = \dot{T}(\balpha, \dot{\balpha}, t) + \dot{V}(\balpha, t) = -2\eta(t) T(\balpha, \dot{\balpha}, t) \leq 0
\end{align}
revealing that the total energy $H(t)$ is monotonically non-increasing in $t$, and hence $0\leq H(t) \leq H(t_0)$ for all $t\geq t_0$. Assuming $H(t_0)<\infty$, the Hamiltonian is bounded below and monotone, and therefore converges to the limit $H_\infty :=\lim_{t\rightarrow \infty} H(t) \in \RR$ by completeness of $\RR$, such that $0\leq H_\infty \leq H(t) \leq H(t_0)$ for all $t\geq t_0$.

\paragraph{Theorem \ref{theorem:lasalle-convergence} - convergence guarantee.}
From Lemma \ref{lemma:energy-dissipation} and under the dynamics in Equation \eqref{eq:dissipative-dynamics-appendix}, recall that $0\leq H_\infty \leq H(t)\leq H(t_0)$ for all $t\geq t_0$, so all trajectories remain in the sublevel set $\Omega=\{(\boldsymbol{\alpha}, \dot{\boldsymbol{\alpha}}) : H(\balpha,\dot{\balpha},t) \leq H(t_0)\}$. If $V(\balpha, t)$ is coercive, i.e. $V(\balpha, t) \rightarrow \infty$ as $\lVert \balpha\rVert \rightarrow \infty$, then since $T(\balpha, \dot{\balpha}, t)\rightarrow\infty$ as $\lVert \dot{\balpha}\rVert\rightarrow \infty$, it holds that
\begin{equation}
    H(\balpha, \dot{\balpha}, t)= T(\balpha, \dot{\balpha}, t) + V(\balpha, t) \rightarrow\infty \quad \textrm{as} \quad \lVert\balpha\rVert \rightarrow\infty \textrm{ or } \lVert\dot{\balpha}\rVert \rightarrow \infty,
\end{equation}
so $H$ is a radially unbounded function in $\balpha$ and $\dot{\balpha}$, and thus $\Omega$ is compact in $\balpha$ and $\dot{\balpha}$.
Moreover assuming that $H$ is continuously differentiable and since $\dot{H}(\balpha, \dot{\balpha},t)\leq 0$ on $\Omega$, the conditions of LaSalle's invariance principle (e.g. in the textbook by \citet{khalil2002nonlinear}) are satisfied and all trajectories converge to the largest invariant subset contained in
\begin{equation}
    \mathcal{S} = \{\left(\balpha, \dot{\balpha}\right):\dot{H}(t) =0\}.
\end{equation}
Recall that $\dot{H}(\balpha, \dot{\balpha}, t) = -2\eta(t) T(\balpha, \dot{\balpha}, t) = -\eta(t) \dot{\balpha}^\top \G{\balpha} \dot{\balpha}\leq 0$, due to positive definiteness of $\G{\balpha}$ and $\eta(t) > 0$ for all $t$, we note that $\dot{H}(t)=0$ if and only if $\dot{\balpha}=\boldsymbol{0}$, hence:
\begin{equation}
    \mathcal{S} = \{\left(\balpha, \dot{\balpha}\right): \dot{\balpha} = \boldsymbol{0}\}.
\end{equation}
Next, note that a trajectory starting in $\mathcal{S}$ remains in $\mathcal{S}$ only if $\ddot{\balpha}=0$. Following the dynamics from Equation \eqref{eq:dissipative-dynamics-appendix} for the particular states $(\balpha, \dot{\balpha})\in\mathcal{S}$, see that:
\begin{align}
    \ddot{\balpha} = - \left(\dot{\balpha}^\top \mathbf{H}_{\mathcal{L}}(\balpha)\dot{\balpha} + \kappa \right)\cdot\operatorname{grad}\Loss{\balpha} - \eta(t) \cdot \dot{\balpha} \overset{(\balpha, \dot{\balpha}) \in \mathcal{S}}{=} -\kappa \operatorname{grad} \Loss{\balpha} = \boldsymbol{0}.
\end{align}
The equality is satisfied if and only if $\operatorname{grad}\Loss{\balpha} = \boldsymbol{0}$ since $\kappa > 0$ by definition, hence the largest invariant subset $\mathcal{I}$ of $\mathcal{S}$ to which all trajectories converge as $t\rightarrow \infty$ is exactly zero-velocity states reaching stationary points of the loss surface: 
\begin{equation}
    \mathcal{I} = \left\{\left(\balpha, \dot{\balpha}\right):\dot{\balpha}= \boldsymbol{0}, \operatorname{grad} \Loss{\balpha} = \boldsymbol{0}\right\} \subset \mathcal{S}.
\end{equation}

Assuming that the loss $\mathcal{L}$ satisfies the strict saddle property, saddle points and local maxima in $\mathcal{I}$ are unstable equilibria and trajectories converge almost surely to a local minimum of $\mathcal{L}$. We thereby conclude that the proposed speed-dependent dissipative dynamics converge to a local minimum of $\mathcal{L}$.

\begin{remark}[Differentiability and coercivity of $V$ in neural network settings.] 
For general neural network loss landscapes, flat or unbounded directions in parameter space may prevent sublevel sets from being compact, as $V$ is not guaranteed to be coercive. In a Bayesian setting with a shifted negative log-joint as the potential, the potential takes the form
$
    V(\balpha) = \kappa \cdot \left(-\log p(\mathcal{D}|\balpha) - \log p(\balpha) - \mathcal{L}(\btheta^\ast_{\text{glob}})\right)
$
where for a Gaussian prior, $p(\balpha) = \mathcal{N}(\boldsymbol{0}, \sigma^2 \mathbb{I}_K)$, the negative log-prior contributes as an $L^2$-regulartization term by $\frac{1}{2\sigma^2}\lVert \balpha\rVert^2$, which grows unboundedly as $\lVert \balpha\rVert \rightarrow \infty$, and the potential of the log-posterior is coercive, which guarantees compactness of $\Omega$.  If moreover the neural network architecture induces a smooth loss function through the activation functions, LaSalle's invariance principle applies.
\end{remark}

\paragraph{Theorem \ref{theorem:convergence-time} - convergence time.}
Under the assumptions stated in Lemma \ref{lemma:energy-dissipation}, the Hamiltonian dissipates energy, hence  $0 \leq H_\infty \leq H(t_1)\leq H(t_0)$. We formulate the energy difference equation of the Hamiltonian as the difference in total energy between the start- and endpoint of the curve as:
\begin{equation}
    \Delta H = H(t_0)-H(t_1) = -\int_{t_0}^{t_1} \dot{H}(t) dt = 2\int_{t_0}^{t_1} \eta(t) T(\balpha, \dot{\balpha},t)dt = 2\eta_0 \int_{t_0}^{t_1} T(\balpha, \dot{\balpha},t)^{3/2}dt,
\end{equation}
Now, let $t_\epsilon$ be a proxy convergence time at which the kinetic energy drops below a threshold $\epsilon$, i.e.
$
    t_{\epsilon} = \inf \left\{t \geq t_0: T(\balpha, \dot{\balpha}, t) \leq \epsilon\right \}.
$
Due to the infimum condition that defines $t_{\epsilon}$ it holds that:
\begin{equation}
    \epsilon^{3/2} (t_{\epsilon} - t_0) = \int_{t_0}^{t_{\epsilon}} \epsilon^{3/2} dt \leq \int_{t_0}^{t_{\epsilon}} T(\balpha, \dot{\balpha}, t)^{3/2} dt,
\end{equation}
Now inserting this lower bound into the difference equation along with boundedness of the Hamiltonian gives:
\begin{equation}
    2\eta_0 \epsilon^{3/2} (t_{\epsilon} - t_0) \leq 2\eta_0 \int_{t_0}^{t_1} T(\balpha, \dot{\balpha},t)^{3/2}dt=\Delta H \leq H(t_0).
\end{equation}
By isolating the proxy convergence time, we get an upper bound on the form:
\begin{equation}
    t_{\epsilon} \leq t_0+\frac{H(t_0)}{2\eta_0\epsilon^{3/2}}.
\end{equation}

\paragraph{Theorem \ref{theorem:gamma-lower-bound} - $\bgamma$-arc-length.}

Given the Hamiltonian described in Lemma \ref{lemma:energy-dissipation} and the formulation of the Hamiltonian difference equation described in Theorem \ref{theorem:convergence-time} along with the fact that $T(\balpha, \dot{\balpha}, t) \leq H(t) \leq H(t_0)$ for any $t \geq t_0$, it holds that:
\begin{equation}
    \Delta H = 2\eta_0\int_{t_0}^{t_1} T(\balpha, \dot{\balpha})^{3/2}dt = 2\eta_0\int_{t_0}^{t_1} \sqrt{T(\balpha, \dot{\balpha})} \cdot T(\balpha, \dot{\balpha})dt \leq 2\eta_0 H(t_0) \int_{t_0}^{t_1} \sqrt{T(\balpha, \dot{\balpha})}dt.
\end{equation}
Note furthermore that the integral of the square-root of the kinetic energy is proportional to the arc-length of the curve $\bgamma:[t_0\rightarrow t_1]\in \M$ by:
\begin{equation}
    \int_{t_0}^{t_1} \sqrt{T(\balpha, \dot{\balpha}})dt = \frac{1}{\sqrt{2}} \int_{t_0}^{t_1} \sqrt{\dot{\balpha}^\top \G{\balpha} \dot{\balpha}} dt = \frac{1}{\sqrt{2}} \int_{t_0}^{t_1} \lVert \dot{\bgamma}\rVert dt = \frac{\textsc{Length}(\bgamma)}{\sqrt{2}}, 
\end{equation}
where $\textsc{Length}(\bgamma)$ denotes the arc-length of the curve. In combination with the upper bound on the energy difference, we get a lower bound on the arc-length:
\begin{equation}
    \frac{\sqrt{2}\Delta H}{2\eta_0 H(t_0)} = \frac{\Delta H}{\sqrt{2} \eta_0 H(t_0)}\leq \textsc{Length}(\bgamma).
\end{equation}
From the convergence properties in Theorem \ref{theorem:gamma-lower-bound}, the asymptotic behavior of the Hamiltonian guarantees that:
\begin{equation}
    \lim_{t\rightarrow\infty} H(t) = \lim_{t\rightarrow\infty} T(\balpha, \dot{\balpha}, t) + \lim_{t\rightarrow\infty} V(\balpha, t) = V(\balpha, \infty) = \kappa \cdot \left(\Loss{\balpha_\infty} - \Loss{\balpha^\ast}\right) \geq 0,
\end{equation}
since $\lim_{t\rightarrow\infty} T(\balpha, \dot{\balpha},t) = 0$. Note that we by $\balpha_\infty$ denote the position to which the trajectory converges where $\operatorname{grad}\Loss{\balpha_\infty} = \boldsymbol{0}$ and remark that $\balpha_\infty$ is not necessarily a global minimum, hence $\Loss{\balpha_\infty} \geq \Loss{\balpha^\ast}$. 
Now, letting the stop time $t_1 \rightarrow \infty$, we see that: 
\begin{align}
    \lim_{t_1\rightarrow \infty} \Delta H & = H(t_0) - \lim_{t_1\rightarrow\infty} H(t_1) \\
    &= T(\balpha, \dot{\balpha}, t_0) + \kappa \cdot \left(\Loss{\balpha(t_0)} - \Loss{\balpha_\infty}\right) \\ 
    &= T(\balpha, \dot{\balpha}, t_0) + \kappa \Delta \mathcal{L}.
\end{align}
Collecting the pieces concludes the proof and we get a lower bound on the form:
\begin{equation}
    \frac{T(\balpha, \dot{\balpha}, t_0) + \kappa\Delta \mathcal{L}}{\sqrt{2}\eta_0 H(t_0)} \leq \textsc{Length}(\bgamma).
\end{equation}

\newpage
\section{Experimental details}
\renewcommand{\theequation}{D.\arabic{equation}}
\setcounter{equation}{0} 

\label{appendix:experimental-details}

\paragraph{Implementation details.} 
All experiments are implemented in PyTorch. Differential equations are solved using the Dormand-Prince 5th-order solver \citep{dormand1980family} from \texttt{torchdiffeq} \citep{chen2018neural}, with absolute and relative tolerances of \texttt{atol=1e-6} and \texttt{rtol=1e-7} to ensure stable dynamics. These tolerances are a computational bottleneck for both standard geodesic sampling (\textsc{RLA}) and for \DIMS at high exploration rates (low $\eta_0$ and large $t_1$), as the adaptive step-size controller requires many evaluations to satisfy the error bounds. Automatic differentiation is used throughout to compute acceleration terms efficiently via Hessian-vector products. For all experiments, the prior precision and noise variance are optimized by maximizing the log-marginal likelihood following the suggestion by \citet{immer2021scalable}. The gravitational pull is fixed to $\kappa=1$ across all runs, leaving the friction coefficient $\eta_0$ as the primary hyperparameter controlling exploration. Networks are trained using the Adam optimizer with learning rate $0.001$ and weight decay $0.01$.

The linearized Laplace (\textsc{LinLA}) uses the same Gaussian samples as the standard, sampled Laplace, i.e. a Gaussian $\btheta_s \sim \mathcal{N}(\btheta^\ast, \mathbf{\Sigma})$ centered at the MAP $\btheta^\ast$ , but it computes predictions via the first-order Taylor expansion $f_{\btheta^\ast}(\bx) + \J{f_{\btheta^\ast}}{\bx}(\btheta_s - \btheta^\ast)$, which can be evaluated efficiently with forward-mode automatic differentiation. \textsc{RLA} uses the centered Laplace samples $\tbv_s = \btheta_s - \btheta^\ast$ as initial velocities and integrates the geodesic equation on the loss manifold from $\btheta^\ast$ for time $t_1$.

For further implementation details, see the code base: \href{https://github.com/albertkjoller/geometric-ml/tree/main/dissipative-riemannian-mechanics}{\texttt{github.com/albertkjoller/geometric-ml}}.

\subsection{Snelson regression} 
\label{appendix:snelson}

A $2$-layer fully connected network with $16$ hidden units per layer and \texttt{GELU} activations was trained for $50,000$ epochs using full-batch gradient descent with $20\%$ dropout. The Laplace approximation was computed using the exact full Hessian over all parameters, with prior precision and variance optimized accordingly. Out-of-distribution data consists of the $52$ points in $x\in[1.5,3]$, and the remaining $N=148$ points form the training set. We set $\eta_0=0.1$ and integrate the dynamics for time $t_1=50$, drawing $S=100$ samples from each method.

\subsection{Banana classification}
\label{appendix:banana-classification}
A $4$-layer fully connected network with $16$ hidden units per layer and \texttt{SiLU} activations was trained for $5,000$ epochs using full-batch gradient descent, with $50\%$ dropout and label smoothing of $0.05$. Of all the available data, we randomly selected $N=265$ points as the training set while the remaining approximately $5,000$ points serve as the test set. We set $\eta_0=0.01$ and integrate for time $t_1=100$, drawing $S=50$ samples from each method.

In Table \ref{tab:banana-classification} we report in-distribution performance on the banana classification task. The \textsc{MAP} estimate achieves the highest accuracy (88.43\%) but is a point estimate and thus provides no valid uncertainty quantification out-of-domain. Both \textsc{LA} and \textsc{LinLA} are too crude approximations, resulting in low accuracy and poor calibration, highlighting the sensitivity of linear Laplace methods to the choice of Hessian approximation in this setting. \textsc{RLA} achieves strong performance across all metrics, closely approaching \textsc{MAP} accuracy (86.49\%) while providing meaningful uncertainty estimates. \DIMS performs comparably to \textsc{RLA} in terms of accuracy (85.49\%), though with slightly higher NLL and calibration errors, and somewhat larger variance across runs — consistent with the higher exploration induced by the dissipative dynamics. Note, however, that this need not always be the case, as evidenced by the remaining experiments where these methods remain competitive in-distribution.

\begin{table}[h!]
  \centering
  \caption{In-distribution performance for the banana classification dataset. We report the mean values over the $S=50$ samples along with the standard error of the mean as the uncertainty estimate. Calibration metrics are computed with $10$ bins.}
  \label{tab:banana-classification}
  \resizebox{1.0\linewidth}{!}{%
  \begin{tabular}{lccccccc}
    \toprule
     \textsc{Method} & Accuracy $\uparrow$ & \textsc{NLL} $\downarrow$ & \textsc{Brier} $\downarrow$ & \textsc{ECE} $\downarrow$ & \textsc{MCE} $\downarrow$ \\
    \midrule
        \textsc{MAP} & $88.43 \pm 0.00$ & $0.275 \pm 0.000$ & $0.0845 \pm 0.0000$ & $1.54 \pm 0.00$ & $7.72 \pm 0.00$ \\ 
        \textsc{LA} & $49.40 \pm 7.07$ & $941.201 \pm 613.928$ & $0.5056 \pm 0.0706$ & $50.58 \pm 7.06$ & $61.61 \pm 15.95$  \\
        \textsc{LinLA} & $52.47 \pm 8.91$ & $57.240 \pm 20.186$ & $0.4720 \pm 0.0888$ & $47.19 \pm 8.87$ & $62.52 \pm 11.80$ \\
        \textsc{RLA} & $86.49 \pm 0.74$ & $0.351 \pm 0.021$ & $0.1025 \pm 0.0045$ & $5.54 \pm 0.85$ & $17.35 \pm 4.18$ \\
        \DIMS (\textsc{Ours}) & $85.49 \pm 1.22$ & $0.412 \pm 0.043$ & $0.1148 \pm 0.0091$ & $7.79 \pm 1.08$ & $25.29 \pm 9.65$ \\
        \bottomrule
  \end{tabular}%
  }
\end{table}

\subsection{UCI classification}
\label{appendix:uci-classification}

A $3$-layer fully connected network with $16$ hidden units per layer and \texttt{SiLU} activations was trained for $10,000$ epochs with a batch size of $64$ and $50\%$ dropout. We draw $S=30$ samples from each method and repeat experiments over $5$ random seeds.

We provide test performance in Table \ref{tab:uci_test} and in-distribution training performance in Table \ref{tab:uci_train}. For each seed, performance metrics are computed by averaging over the $S=30$ sample, and results are reported as the mean with the standard error across the 5 seeds as the uncertainty estimate. Calibration errors are computed using $10$ bins. Overall, \DIMS performs competitively with \textsc{RLA} across datasets, with the geometry-aware methods generally dominating over the standard approximations. As in the banana experiment, sampling-based methods occasionally trade a degree of in-distribution accuracy for improved calibration and uncertainty quantification. This is seen both on the test and training set.

\subsection{MNIST and Fashion MNIST} 
\label{appendix:mnist-classification}

We train a LeNet classifier (approx. $44{,}000$ parameters) on MNIST and Fashion MNIST using Adam for $10$ epochs using a batch size of $128$, with a log-softmax output layer and negative log-likelihood loss. After training, the MAP checkpoint $\btheta^\ast$ serves as the reference point for all posterior approximations.

We compare \DIMS to Laplace-based approximations that fit a Gaussian $\mathcal{N}(\btheta^\ast, \mathbf{\Sigma})$ where $\Sigma^{-1} = \mathbf{H}_\mathcal{L}(\btheta^\ast)$ is the Hessian of the negative log-posterior. For computational reasons, we estimate the Hessian matrix on a randomly selected subset of $N=1{,}000$ training samples and since the full Hessian is intractable for a network with a high-dimensional parameter space, we resort to an approximation of the Hessian. For \DIMS we set the friction coefficient to $\eta_0=0.1$ and consider the study of three Hessian approximations - specifically the diagonal (D), Kronecker-factored Approximate Curvature (KFAC or K) and a low-rank approximation (LR) based on the eigenvalues of the full Hessian - as an ablation study on which approximate Hessian to use for further experiments. As can be seen in Table \ref{tab:hessian-influence}, the diagonal approximation underfits severely in the standard and linearized Laplace settings, in fact to such an extend that the initial velocities results in unstable behavior of the geometry-based methods that never converge within the tolerances set, for which reason we exclude these results. While KFAC also results in underfitting for the standard Laplace and linearized Laplace, the geometry-aware methods perform relatively well on the task with \textsc{DiMS-K} being superior to \textsc{RLA-K}. Lastly, the LR approximations improve results significantly for the sampled, linearized and Riemannian Laplace approximations, and also for \DIMS on certaint metrics. We remark that \DIMS is superior or on par with \textsc{RLA}. 

With this experiment we therefore highlight the limitation of the proposed methods originating from the approximation of the Hessian that is used to generate the initial velocities. Nevertheless, \DIMS appears to be somewhat robust to this choice, which is a clear benefit of the approach. We continue with the low-rank approximation and evaluate the generated samples from each method on out-of-distribution (OOD) data via the AUROC score. For MNIST the OOD sets are FashionMNIST, EMNIST, and KMNIST, and for FashionMNIST they are MNIST, EMNIST, and KMNIST. These results are provided in Table \ref{tab:ood-main} in the main paper. We provide ID and OOD results under different settings in Tables \ref{tab:id-seed0}-\ref{tab:id-seed2}.

\begin{table}[ht]
  \centering
  \caption{Influence of the Hessian approximation. Remark that the diagonal approximation is unstable for the geometric methods, given the already low tolerances, and therefore excluded. Results are based on a single seed for exploration purposes and serves as a first step targeting which approximation to choose for further experiments.}
  \label{tab:hessian-influence}
  \resizebox{\linewidth}{!}{%
  \begin{tabular}{clccccc}
    \toprule
     &  & \textsc{Acc.} $\uparrow$ & \textsc{NLL} $\downarrow$ & \textsc{Brier} $\downarrow$ & \textsc{ECE} $\downarrow$ & \textsc{MCE} $\downarrow$ \\
    \midrule
    \multirow{11}{*}{\rotatebox{90}{\textsc{MNIST}}} & \cellcolor{gray!10}\textsc{Map} & \cellcolor{gray!10}$0.991$ & \cellcolor{gray!10}$0.095$ & \cellcolor{gray!10}$0.002$ & \cellcolor{gray!10}$0.008$ & \cellcolor{gray!10}$0.693$ \\
    \cmidrule(lr){2-7}
     & \textsc{LA-D} & $0.101 \pm 0.003$ & $240052.914 \pm 10136.718$ & $0.180 \pm 0.001$ & $0.899 \pm 0.003$ & $0.954 \pm 0.005$ \\
     & \textsc{LA-K} & $0.099 \pm 0.002$ & $155383.233 \pm 4715.344$ & $0.180 \pm 0.0004$ & $0.901 \pm 0.002$ & $0.958 \pm 0.005$ \\
     & \textsc{LA-LR} & $0.564 \pm 0.009$ & $11.502 \pm 0.507$ & $0.083 \pm 0.002$ & $0.408 \pm 0.009$ & $0.614 \pm 0.006$ \\
    \cmidrule(lr){2-7}
     & \textsc{LinLA-D} & $0.249 \pm 0.012$ & $208.416 \pm 6.798$ & $0.149 \pm 0.002$ & $0.745 \pm 0.012$ & $0.836 \pm 0.006$ \\
     & \textsc{LinLA-K} & $0.268 \pm 0.011$ & $156.960 \pm 4.476$ & $0.145 \pm 0.002$ & $0.725 \pm 0.011$ & $0.818 \pm 0.005$ \\
     & \textsc{LinLA-LR} & $0.860 \pm 0.007$ & $2.608 \pm 0.151$ & $0.026 \pm 0.001$ & $0.127 \pm 0.006$ & $0.508 \pm 0.007$ \\
    \cmidrule(lr){2-7}
     & \textsc{RLA-K} & $0.908 \pm 0.001$ & $1.564 \pm 0.054$ & $0.017 \pm 0.0002$ & $0.083 \pm 0.001$ & $0.482 \pm 0.008$ \\
     & \textsc{RLA-LR} & \cellcolor{gray!15}$\mathbf{0.953 \pm 0.001}$ & $0.609 \pm 0.009$ & $0.009 \pm 0.0001$ & $0.042 \pm 0.0005$ & $0.472 \pm 0.008$ \\
    \cmidrule(lr){2-7}
     & \textsc{DiMS-K} & $0.946 \pm 0.0005$ & \cellcolor{gray!15}$\mathbf{0.232 \pm 0.003}$ & $0.009 \pm 8e-05$ & \cellcolor{gray!15}$\mathbf{0.030 \pm 0.0003}$ & \cellcolor{gray!15}$\mathbf{0.305 \pm 0.014}$ \\
     & \textsc{DiMS-LR} & $0.950 \pm 0.0002$ & $0.271 \pm 0.002$ & \cellcolor{gray!15}$\mathbf{0.008 \pm 4e-05}$ & $0.035 \pm 0.0002$ & $0.379 \pm 0.009$ \\
    \bottomrule
  \end{tabular}%
  }
\end{table}

\begin{table}[ht]
  \centering
  \caption{UCI results on the test sets. We report the mean of 5 seeds along with the standard error.}
  \label{tab:uci_test}
  \resizebox{\linewidth}{!}{%
  \begin{tabular}{clcccccc}
    \toprule
     &  & \textsc{Acc.} $\uparrow$ & \textsc{NLL} $\downarrow$ & \textsc{Brier} $\downarrow$ & \textsc{ECE} $\downarrow$ & \textsc{MCE} $\downarrow$ & \textsc{AUROC} $\uparrow$ \\
    \midrule
    \multirow{6}{*}{\rotatebox{90}{\textsc{Vehicle}}} & \textsc{Map} & $0.733 \pm 0.010$ & $0.485 \pm 0.007$ & $0.302 \pm 0.004$ & $0.061 \pm 0.004$ & $0.286 \pm 0.030$ & $0.917 \pm 0.002$ \\
     & \textsc{LA} & $0.260 \pm 0.004$ & $17.852 \pm 0.175$ & $1.456 \pm 0.007$ & $0.725 \pm 0.003$ & $0.824 \pm 0.002$ & $0.521 \pm 0.003$ \\
     & \textsc{LinLA} & $0.443 \pm 0.008$ & $5.446 \pm 0.327$ & $0.916 \pm 0.017$ & $0.401 \pm 0.010$ & $0.617 \pm 0.011$ & $0.664 \pm 0.011$ \\
     & \begin{tabular}[c]{@{}l@{}}\textsc{RLA} \\ {\footnotesize $t=1.0$}\end{tabular} & $0.809 \pm 0.009$ & $0.776 \pm 0.051$ & $0.302 \pm 0.015$ & $0.130 \pm 0.007$ & $0.401 \pm 0.006$ & $0.950 \pm 0.003$ \\
    \cmidrule(lr){2-8}
     & \begin{tabular}[c]{@{}l@{}}\textsc{DiMS} \\ {\footnotesize $t=50.0, \eta_0=0.1$}\end{tabular} & $0.819 \pm 0.018$ & $0.791 \pm 0.077$ & $0.299 \pm 0.030$ & $0.136 \pm 0.015$ & $0.462 \pm 0.026$ & $0.955 \pm 0.004$ \\
     & \begin{tabular}[c]{@{}l@{}}\textsc{DiMS} \\ {\footnotesize $t=50.0, \eta_0=0.5$}\end{tabular} & $0.778 \pm 0.011$ & $0.422 \pm 0.015$ & $0.272 \pm 0.009$ & $0.061 \pm 0.005$ & $0.385 \pm 0.011$ & $0.935 \pm 0.003$ \\
    \midrule
    \multirow{6}{*}{\rotatebox{90}{\textsc{Glass}}} & \textsc{Map} & $0.591 \pm 0.016$ & $1.070 \pm 0.052$ & $0.552 \pm 0.010$ & $0.100 \pm 0.014$ & $0.511 \pm 0.112$ & $0.829 \pm 0.022$ \\
     & \textsc{LA} & $0.165 \pm 0.005$ & $20.855 \pm 0.299$ & $1.642 \pm 0.010$ & $0.818 \pm 0.006$ & $0.903 \pm 0.005$ & $0.476 \pm 0.008$ \\
     & \textsc{LinLA} & $0.302 \pm 0.014$ & $4.097 \pm 0.233$ & $1.063 \pm 0.025$ & $0.452 \pm 0.017$ & $0.835 \pm 0.022$ & $0.599 \pm 0.005$ \\
     & \begin{tabular}[c]{@{}l@{}}\textsc{RLA} \\ {\footnotesize $t=1.0$}\end{tabular} & $0.683 \pm 0.014$ & $2.309 \pm 0.192$ & $0.519 \pm 0.023$ & $0.232 \pm 0.013$ & $0.544 \pm 0.024$ & $0.868 \pm 0.021$ \\
    \cmidrule(lr){2-8}
     & \begin{tabular}[c]{@{}l@{}}\textsc{DiMS} \\ {\footnotesize $t=50.0, \eta_0=0.1$}\end{tabular} & $0.714 \pm 0.023$ & $1.749 \pm 0.312$ & $0.473 \pm 0.042$ & $0.214 \pm 0.023$ & $0.572 \pm 0.015$ & $0.872 \pm 0.033$ \\
     & \begin{tabular}[c]{@{}l@{}}\textsc{DiMS} \\ {\footnotesize $t=50.0, \eta_0=0.5$}\end{tabular} & $0.733 \pm 0.017$ & $1.317 \pm 0.190$ & $0.436 \pm 0.027$ & $0.179 \pm 0.012$ & $0.504 \pm 0.013$ & $0.901 \pm 0.022$ \\
    \midrule
    \multirow{6}{*}{\rotatebox{90}{\textsc{Ionosphere}}} & \textsc{Map} & $0.907 \pm 0.010$ & $0.234 \pm 0.018$ & $0.140 \pm 0.011$ & $0.055 \pm 0.003$ & $0.323 \pm 0.040$ & $0.967 \pm 0.008$ \\
     & \textsc{LA} & $0.462 \pm 0.022$ & $11.857 \pm 0.691$ & $1.061 \pm 0.044$ & $0.534 \pm 0.022$ & $0.693 \pm 0.015$ & $0.585 \pm 0.027$ \\
     & \textsc{LinLA} & $0.697 \pm 0.013$ & $2.591 \pm 0.179$ & $0.549 \pm 0.025$ & $0.274 \pm 0.012$ & $0.605 \pm 0.008$ & $0.702 \pm 0.020$ \\
     & \begin{tabular}[c]{@{}l@{}}\textsc{RLA} \\ {\footnotesize $t=1.0$}\end{tabular} & $0.900 \pm 0.004$ & $0.634 \pm 0.066$ & $0.177 \pm 0.006$ & $0.091 \pm 0.003$ & $0.550 \pm 0.026$ & $0.925 \pm 0.006$ \\
    \cmidrule(lr){2-8}
     & \begin{tabular}[c]{@{}l@{}}\textsc{DiMS} \\ {\footnotesize $t=50.0, \eta_0=0.1$}\end{tabular} & $0.892 \pm 0.012$ & $0.608 \pm 0.051$ & $0.195 \pm 0.016$ & $0.098 \pm 0.008$ & $0.522 \pm 0.071$ & $0.902 \pm 0.015$ \\
     & \begin{tabular}[c]{@{}l@{}}\textsc{DiMS} \\ {\footnotesize $t=50.0, \eta_0=0.5$}\end{tabular} & $0.884 \pm 0.010$ & $0.724 \pm 0.092$ & $0.210 \pm 0.015$ & $0.103 \pm 0.007$ & $0.530 \pm 0.051$ & $0.881 \pm 0.018$ \\
    \midrule
    \multirow{6}{*}{\rotatebox{90}{\textsc{Waveform}}} & \textsc{Map} & $0.861 \pm 0.002$ & $0.315 \pm 0.006$ & $0.193 \pm 0.003$ & $0.040 \pm 0.002$ & $0.294 \pm 0.113$ & $0.972 \pm 0.001$ \\
     & \textsc{LA} & $0.443 \pm 0.011$ & $4.518 \pm 0.218$ & $0.982 \pm 0.020$ & $0.462 \pm 0.010$ & $0.610 \pm 0.016$ & $0.640 \pm 0.012$ \\
     & \textsc{LinLA} & $0.614 \pm 0.010$ & $1.287 \pm 0.049$ & $0.605 \pm 0.016$ & $0.243 \pm 0.009$ & $0.417 \pm 0.018$ & $0.811 \pm 0.008$ \\
     & \begin{tabular}[c]{@{}l@{}}\textsc{RLA} \\ {\footnotesize $t=1.0$}\end{tabular} & $0.840 \pm 0.001$ & $0.416 \pm 0.015$ & $0.228 \pm 0.003$ & $0.063 \pm 0.002$ & $0.193 \pm 0.015$ & $0.964 \pm 0.001$ \\
    \cmidrule(lr){2-8}
     & \begin{tabular}[c]{@{}l@{}}\textsc{DiMS} \\ {\footnotesize $t=50.0, \eta_0=0.1$}\end{tabular} & $0.848 \pm 0.003$ & $0.369 \pm 0.021$ & $0.212 \pm 0.005$ & $0.052 \pm 0.006$ & $0.185 \pm 0.011$ & $0.968 \pm 0.002$ \\
     & \begin{tabular}[c]{@{}l@{}}\textsc{DiMS} \\ {\footnotesize $t=50.0, \eta_0=0.5$}\end{tabular} & $0.863 \pm 0.001$ & $0.299 \pm 0.007$ & $0.189 \pm 0.003$ & $0.020 \pm 0.001$ & $0.227 \pm 0.039$ & $0.972 \pm 0.001$ \\
    \midrule
    \multirow{6}{*}{\rotatebox{90}{\textsc{Australian}}} & \textsc{Map} & $0.865 \pm 0.008$ & $0.314 \pm 0.020$ & $0.185 \pm 0.011$ & $0.053 \pm 0.013$ & $0.249 \pm 0.062$ & $0.939 \pm 0.009$ \\
     & \textsc{LA} & $0.535 \pm 0.008$ & $6.577 \pm 0.231$ & $0.881 \pm 0.016$ & $0.438 \pm 0.008$ & $0.613 \pm 0.008$ & $0.545 \pm 0.022$ \\
     & \textsc{LinLA} & $0.724 \pm 0.007$ & $0.789 \pm 0.020$ & $0.420 \pm 0.010$ & $0.168 \pm 0.003$ & $0.368 \pm 0.014$ & $0.793 \pm 0.009$ \\
     & \begin{tabular}[c]{@{}l@{}}\textsc{RLA} \\ {\footnotesize $t=1.0$}\end{tabular} & $0.843 \pm 0.009$ & $0.712 \pm 0.093$ & $0.257 \pm 0.018$ & $0.107 \pm 0.011$ & $0.366 \pm 0.034$ & $0.902 \pm 0.011$ \\
    \cmidrule(lr){2-8}
     & \begin{tabular}[c]{@{}l@{}}\textsc{DiMS} \\ {\footnotesize $t=50.0, \eta_0=0.1$}\end{tabular} & $0.810 \pm 0.008$ & $1.168 \pm 0.075$ & $0.335 \pm 0.015$ & $0.164 \pm 0.007$ & $0.498 \pm 0.017$ & $0.876 \pm 0.009$ \\
     & \begin{tabular}[c]{@{}l@{}}\textsc{DiMS} \\ {\footnotesize $t=50.0, \eta_0=0.5$}\end{tabular} & $0.816 \pm 0.009$ & $1.258 \pm 0.109$ & $0.325 \pm 0.018$ & $0.155 \pm 0.010$ & $0.487 \pm 0.017$ & $0.871 \pm 0.012$ \\
    \midrule
    \multirow{6}{*}{\rotatebox{90}{\textsc{Breast C.}}} & \textsc{Map} & $0.981 \pm 0.003$ & $0.062 \pm 0.006$ & $0.035 \pm 0.004$ & $0.021 \pm 0.002$ & $0.421 \pm 0.093$ & $0.997 \pm 0.001$ \\
     & \textsc{LA} & $0.553 \pm 0.019$ & $10.084 \pm 0.479$ & $0.880 \pm 0.038$ & $0.442 \pm 0.019$ & $0.657 \pm 0.008$ & $0.551 \pm 0.014$ \\
     & \textsc{LinLA} & $0.745 \pm 0.007$ & $1.662 \pm 0.079$ & $0.447 \pm 0.013$ & $0.216 \pm 0.007$ & $0.541 \pm 0.010$ & $0.819 \pm 0.014$ \\
     & \begin{tabular}[c]{@{}l@{}}\textsc{RLA} \\ {\footnotesize $t=1.0$}\end{tabular} & $0.969 \pm 0.003$ & $0.185 \pm 0.016$ & $0.055 \pm 0.006$ & $0.030 \pm 0.003$ & $0.514 \pm 0.017$ & $0.992 \pm 0.001$ \\
    \cmidrule(lr){2-8}
     & \begin{tabular}[c]{@{}l@{}}\textsc{DiMS} \\ {\footnotesize $t=50.0, \eta_0=0.1$}\end{tabular} & $0.969 \pm 0.008$ & $0.123 \pm 0.012$ & $0.055 \pm 0.011$ & $0.029 \pm 0.006$ & $0.501 \pm 0.126$ & $0.995 \pm 0.001$ \\
     & \begin{tabular}[c]{@{}l@{}}\textsc{DiMS} \\ {\footnotesize $t=50.0, \eta_0=0.5$}\end{tabular} & $0.970 \pm 0.005$ & $0.117 \pm 0.013$ & $0.051 \pm 0.008$ & $0.028 \pm 0.005$ & $0.548 \pm 0.083$ & $0.995 \pm 0.0004$ \\
    \bottomrule
  \end{tabular}%
  }
\end{table}

\begin{table}[ht]
  \centering
  \caption{UCI results on the training sets. We report the mean of 5 seeds along with the standard error.}
  \label{tab:uci_train}
  \resizebox{\linewidth}{!}{%
  \begin{tabular}{clcccccc}
    \toprule
     &  & \textsc{Acc.} $\uparrow$ & \textsc{NLL} $\downarrow$ & \textsc{Brier} $\downarrow$ & \textsc{ECE} $\downarrow$ & \textsc{MCE} $\downarrow$ & \textsc{AUROC} $\uparrow$ \\
    \midrule
    \multirow{6}{*}{\rotatebox{90}{\textsc{Vehicle}}} & \textsc{Map} & $0.758 \pm 0.005$ & $0.465 \pm 0.004$ & $0.294 \pm 0.001$ & $0.071 \pm 0.005$ & $0.337 \pm 0.096$ & $0.925 \pm 0.002$ \\
     & \textsc{LA} & $0.259 \pm 0.003$ & $17.891 \pm 0.221$ & $1.458 \pm 0.006$ & $0.725 \pm 0.003$ & $0.786 \pm 0.005$ & $0.519 \pm 0.002$ \\
     & \textsc{LinLA} & $0.438 \pm 0.005$ & $5.445 \pm 0.354$ & $0.920 \pm 0.015$ & $0.402 \pm 0.010$ & $0.582 \pm 0.007$ & $0.663 \pm 0.011$ \\
     & \begin{tabular}[c]{@{}l@{}}\textsc{RLA} \\ {\footnotesize $t=1.0$}\end{tabular} & $0.932 \pm 0.001$ & $0.150 \pm 0.003$ & $0.095 \pm 0.002$ & $0.016 \pm 0.0004$ & $0.279 \pm 0.018$ & $0.991 \pm 0.0005$ \\
    \cmidrule(lr){2-8}
     & \begin{tabular}[c]{@{}l@{}}\textsc{DiMS} \\ {\footnotesize $t=50.0, \eta_0=0.1$}\end{tabular} & $0.989 \pm 0.004$ & $0.047 \pm 0.008$ & $0.022 \pm 0.005$ & $0.024 \pm 0.001$ & $0.347 \pm 0.009$ & $0.999 \pm 0.001$ \\
     & \begin{tabular}[c]{@{}l@{}}\textsc{DiMS} \\ {\footnotesize $t=50.0, \eta_0=0.5$}\end{tabular} & $0.847 \pm 0.003$ & $0.297 \pm 0.005$ & $0.198 \pm 0.004$ & $0.029 \pm 0.001$ & $0.325 \pm 0.022$ & $0.962 \pm 0.001$ \\
    \midrule
    \multirow{6}{*}{\rotatebox{90}{\textsc{Glass}}} & \textsc{Map} & $0.643 \pm 0.007$ & $0.885 \pm 0.013$ & $0.498 \pm 0.007$ & $0.101 \pm 0.008$ & $0.341 \pm 0.096$ & $0.887 \pm 0.003$ \\
     & \textsc{LA} & $0.163 \pm 0.003$ & $20.756 \pm 0.324$ & $1.643 \pm 0.008$ & $0.816 \pm 0.004$ & $0.892 \pm 0.005$ & $0.473 \pm 0.004$ \\
     & \textsc{LinLA} & $0.318 \pm 0.013$ & $3.844 \pm 0.286$ & $1.047 \pm 0.032$ & $0.432 \pm 0.024$ & $0.792 \pm 0.009$ & $0.621 \pm 0.011$ \\
     & \begin{tabular}[c]{@{}l@{}}\textsc{RLA} \\ {\footnotesize $t=1.0$}\end{tabular} & $0.900 \pm 0.004$ & $0.242 \pm 0.011$ & $0.142 \pm 0.005$ & $0.042 \pm 0.001$ & $0.368 \pm 0.013$ & $0.990 \pm 0.001$ \\
    \cmidrule(lr){2-8}
     & \begin{tabular}[c]{@{}l@{}}\textsc{DiMS} \\ {\footnotesize $t=50.0, \eta_0=0.1$}\end{tabular} & $0.993 \pm 0.002$ & $0.066 \pm 0.006$ & $0.023 \pm 0.004$ & $0.050 \pm 0.003$ & $0.398 \pm 0.004$ & $1.000 \pm 5e-05$ \\
     & \begin{tabular}[c]{@{}l@{}}\textsc{DiMS} \\ {\footnotesize $t=50.0, \eta_0=0.5$}\end{tabular} & $0.927 \pm 0.002$ & $0.210 \pm 0.006$ & $0.112 \pm 0.003$ & $0.059 \pm 0.002$ & $0.380 \pm 0.015$ & $0.994 \pm 0.0004$ \\
    \midrule
    \multirow{6}{*}{\rotatebox{90}{\textsc{Ionosphere}}} & \textsc{Map} & $0.982 \pm 0.001$ & $0.059 \pm 0.004$ & $0.028 \pm 0.002$ & $0.024 \pm 0.002$ & $0.367 \pm 0.095$ & $0.998 \pm 0.0004$ \\
     & \textsc{LA} & $0.471 \pm 0.022$ & $11.521 \pm 0.637$ & $1.040 \pm 0.045$ & $0.520 \pm 0.023$ & $0.677 \pm 0.015$ & $0.593 \pm 0.026$ \\
     & \textsc{LinLA} & $0.715 \pm 0.010$ & $2.310 \pm 0.145$ & $0.508 \pm 0.020$ & $0.241 \pm 0.010$ & $0.449 \pm 0.014$ & $0.742 \pm 0.016$ \\
     & \begin{tabular}[c]{@{}l@{}}\textsc{RLA} \\ {\footnotesize $t=1.0$}\end{tabular} & $0.997 \pm 0.001$ & $0.017 \pm 0.002$ & $0.007 \pm 0.001$ & $0.011 \pm 0.0005$ & $0.381 \pm 0.027$ & $1.000 \pm 5e-05$ \\
    \cmidrule(lr){2-8}
     & \begin{tabular}[c]{@{}l@{}}\textsc{DiMS} \\ {\footnotesize $t=50.0, \eta_0=0.1$}\end{tabular} & $0.998 \pm 0.001$ & $0.010 \pm 0.001$ & $0.002 \pm 0.001$ & $0.009 \pm 0.001$ & $0.431 \pm 0.107$ & $1.000$ \\
     & \begin{tabular}[c]{@{}l@{}}\textsc{DiMS} \\ {\footnotesize $t=50.0, \eta_0=0.5$}\end{tabular} & $0.998 \pm 0.001$ & $0.010 \pm 0.001$ & $0.002 \pm 0.0004$ & $0.009 \pm 0.001$ & $0.454 \pm 0.062$ & $1.000$ \\
    \midrule
    \multirow{6}{*}{\rotatebox{90}{\textsc{Waveform}}} & \textsc{Map} & $0.875 \pm 0.001$ & $0.295 \pm 0.002$ & $0.179 \pm 0.001$ & $0.047 \pm 0.001$ & $0.357 \pm 0.115$ & $0.977 \pm 0.0004$ \\
     & \textsc{LA} & $0.445 \pm 0.010$ & $4.497 \pm 0.202$ & $0.980 \pm 0.019$ & $0.459 \pm 0.009$ & $0.589 \pm 0.015$ & $0.641 \pm 0.012$ \\
     & \textsc{LinLA} & $0.620 \pm 0.011$ & $1.257 \pm 0.049$ & $0.596 \pm 0.018$ & $0.236 \pm 0.010$ & $0.429 \pm 0.030$ & $0.816 \pm 0.008$ \\
     & \begin{tabular}[c]{@{}l@{}}\textsc{RLA} \\ {\footnotesize $t=1.0$}\end{tabular} & $0.900 \pm 0.001$ & $0.210 \pm 0.003$ & $0.136 \pm 0.002$ & $0.009 \pm 0.0002$ & $0.129 \pm 0.011$ & $0.985 \pm 0.0004$ \\
    \cmidrule(lr){2-8}
     & \begin{tabular}[c]{@{}l@{}}\textsc{DiMS} \\ {\footnotesize $t=50.0, \eta_0=0.1$}\end{tabular} & $0.908 \pm 0.003$ & $0.195 \pm 0.007$ & $0.125 \pm 0.005$ & $0.012 \pm 0.0003$ & $0.131 \pm 0.043$ & $0.987 \pm 0.001$ \\
     & \begin{tabular}[c]{@{}l@{}}\textsc{DiMS} \\ {\footnotesize $t=50.0, \eta_0=0.5$}\end{tabular} & $0.879 \pm 0.001$ & $0.264 \pm 0.003$ & $0.167 \pm 0.001$ & $0.015 \pm 0.002$ & $0.275 \pm 0.044$ & $0.978 \pm 0.0004$ \\
    \midrule
    \multirow{6}{*}{\rotatebox{90}{\textsc{Australian}}} & \textsc{Map} & $0.883 \pm 0.003$ & $0.277 \pm 0.006$ & $0.166 \pm 0.004$ & $0.024 \pm 0.002$ & $0.108 \pm 0.016$ & $0.953 \pm 0.002$ \\
     & \textsc{LA} & $0.530 \pm 0.008$ & $6.687 \pm 0.267$ & $0.894 \pm 0.016$ & $0.439 \pm 0.008$ & $0.549 \pm 0.007$ & $0.537 \pm 0.022$ \\
     & \textsc{LinLA} & $0.726 \pm 0.011$ & $0.755 \pm 0.032$ & $0.414 \pm 0.016$ & $0.154 \pm 0.007$ & $0.261 \pm 0.008$ & $0.800 \pm 0.013$ \\
     & \begin{tabular}[c]{@{}l@{}}\textsc{RLA} \\ {\footnotesize $t=1.0$}\end{tabular} & $0.942 \pm 0.003$ & $0.144 \pm 0.008$ & $0.086 \pm 0.005$ & $0.017 \pm 0.0003$ & $0.139 \pm 0.010$ & $0.987 \pm 0.001$ \\
    \cmidrule(lr){2-8}
     & \begin{tabular}[c]{@{}l@{}}\textsc{DiMS} \\ {\footnotesize $t=50.0, \eta_0=0.1$}\end{tabular} & $0.997 \pm 0.001$ & $0.021 \pm 0.002$ & $0.007 \pm 0.001$ & $0.017 \pm 0.001$ & $0.433 \pm 0.036$ & $1.000 \pm 2e-05$ \\
     & \begin{tabular}[c]{@{}l@{}}\textsc{DiMS} \\ {\footnotesize $t=50.0, \eta_0=0.5$}\end{tabular} & $0.989 \pm 0.002$ & $0.043 \pm 0.005$ & $0.020 \pm 0.003$ & $0.018 \pm 0.0003$ & $0.309 \pm 0.033$ & $0.999 \pm 0.0004$ \\
    \midrule
    \multirow{6}{*}{\rotatebox{90}{\textsc{Breast C.}}} & \textsc{Map} & $0.989 \pm 0.001$ & $0.049 \pm 0.001$ & $0.024 \pm 0.001$ & $0.016 \pm 0.001$ & $0.310 \pm 0.012$ & $0.998 \pm 0.0002$ \\
     & \textsc{LA} & $0.556 \pm 0.018$ & $10.034 \pm 0.443$ & $0.875 \pm 0.036$ & $0.437 \pm 0.018$ & $0.624 \pm 0.013$ & $0.550 \pm 0.014$ \\
     & \textsc{LinLA} & $0.746 \pm 0.005$ & $1.670 \pm 0.081$ & $0.446 \pm 0.011$ & $0.208 \pm 0.006$ & $0.430 \pm 0.009$ & $0.817 \pm 0.014$ \\
     & \begin{tabular}[c]{@{}l@{}}\textsc{RLA} \\ {\footnotesize $t=1.0$}\end{tabular} & $0.998 \pm 0.0004$ & $0.010 \pm 0.001$ & $0.004 \pm 0.0004$ & $0.006 \pm 0.0002$ & $0.412 \pm 0.011$ & $1.000 \pm 3e-06$ \\
    \cmidrule(lr){2-8}
     & \begin{tabular}[c]{@{}l@{}}\textsc{DiMS} \\ {\footnotesize $t=50.0, \eta_0=0.1$}\end{tabular} & $1.000$ & $0.005 \pm 0.0003$ & $0.001 \pm 0.0001$ & $0.004 \pm 0.0003$ & $0.183 \pm 0.024$ & $1.000$ \\
     & \begin{tabular}[c]{@{}l@{}}\textsc{DiMS} \\ {\footnotesize $t=50.0, \eta_0=0.5$}\end{tabular} & $1.000$ & $0.005 \pm 0.0002$ & $0.001 \pm 6e-05$ & $0.004 \pm 0.0001$ & $0.171 \pm 0.016$ & $1.000$ \\
    \bottomrule
  \end{tabular}%
  }
\end{table}

\newpage

\begin{table}[ht]
  \centering
  \caption{ID results for a random seed on MNIST and Fashion MNIST. We show that the running time does not matter much, when compared to the \textsc{RLA} baseline. We use the low-rank Hessian approximation for all cases, and provide the associated OOD detection results below, given by the \textsc{AUROC} score ($\uparrow$).}
  \label{tab:id-seed0}
  \resizebox{\linewidth}{!}{%
  \begin{tabular}{clccccc}
    \toprule
     &  & \textsc{Acc.} $\uparrow$ & \textsc{NLL} $\downarrow$ & \textsc{Brier} $\downarrow$ & \textsc{ECE} $\downarrow$ & \textsc{MCE} $\downarrow$ \\
    \midrule
    \multirow{4}{*}{\rotatebox{90}{\textsc{MNIST}}} & \textsc{Map} & $0.991$ & $0.095$ & $0.002$ & $0.008$ & $0.693$ \\
     & \textsc{LA} & $0.564 \pm 0.009$ & $11.502 \pm 0.507$ & $0.083 \pm 0.002$ & $0.408 \pm 0.009$ & $0.614 \pm 0.006$ \\
     & \textsc{LinLA} & $0.860 \pm 0.007$ & $2.608 \pm 0.151$ & $0.026 \pm 0.001$ & $0.127 \pm 0.006$ & $0.508 \pm 0.007$ \\
     & \textsc{RLA} & $0.953 \pm 0.001$ & $0.609 \pm 0.009$ & $0.009 \pm 0.0001$ & $0.042 \pm 0.0005$ & $0.472 \pm 0.008$ \\
    \cmidrule(lr){2-7}
     & \begin{tabular}[c]{@{}l@{}}\textsc{DiMS} \\ {\footnotesize $t=25.0, \eta_0=0.1$}\end{tabular} & $0.959 \pm 0.0002$ & $0.209 \pm 0.001$ & $0.007 \pm 4e-05$ & $0.028 \pm 0.0002$ & $0.391 \pm 0.018$ \\
     & \begin{tabular}[c]{@{}l@{}}\textsc{DiMS} \\ {\footnotesize $t=50.0, \eta_0=0.1$}\end{tabular} & $0.950 \pm 0.0002$ & $0.271 \pm 0.002$ & $0.008 \pm 4e-05$ & $0.035 \pm 0.0002$ & $0.379 \pm 0.009$ \\
    \midrule
    \multirow{4}{*}{\rotatebox{90}{\textsc{FMNIST}}} & \textsc{Map} & $0.888$ & $0.479$ & $0.018$ & $0.067$ & $0.231$ \\
     & \textsc{LA} & $0.147 \pm 0.003$ & $53.519 \pm 1.298$ & $0.167 \pm 0.001$ & $0.825 \pm 0.003$ & $0.866 \pm 0.003$ \\
     & \textsc{LinLA} & $0.385 \pm 0.009$ & $11.636 \pm 0.270$ & $0.115 \pm 0.002$ & $0.555 \pm 0.008$ & $0.647 \pm 0.006$ \\
     & \textsc{RLA} & $0.767 \pm 0.002$ & $9.210 \pm 1.266$ & $0.045 \pm 0.0004$ & $0.220 \pm 0.002$ & $0.520 \pm 0.006$ \\
    \cmidrule(lr){2-7}
     & \begin{tabular}[c]{@{}l@{}}\textsc{DiMS} \\ {\footnotesize $t=25.0, \eta_0=0.1$}\end{tabular} & $0.814 \pm 0.002$ & $1.845 \pm 0.106$ & $0.033 \pm 0.0004$ & $0.153 \pm 0.002$ & $0.449 \pm 0.012$ \\
     & \begin{tabular}[c]{@{}l@{}}\textsc{DiMS} \\ {\footnotesize $t=50.0, \eta_0=0.1$}\end{tabular} & $0.814 \pm 0.001$ & $1.473 \pm 0.012$ & $0.033 \pm 0.0001$ & $0.149 \pm 0.0005$ & $0.415 \pm 0.007$ \\
    \bottomrule
  \end{tabular}%
  }

  \vspace{20pt}
  \resizebox{\linewidth}{!}{%
  \begin{tabular}{lcccccc}
    \toprule
     Trained on: & \multicolumn{3}{c}{\leavevmode\leaders\hrule height .55ex depth -.45ex\hfill\kern0pt\enspace\textsc{MNIST}\enspace\leavevmode\leaders\hrule height .55ex depth -.45ex\hfill\kern0pt} & \multicolumn{3}{c}{\leavevmode\leaders\hrule height .55ex depth -.45ex\hfill\kern0pt\enspace\textsc{FMNIST}\enspace\leavevmode\leaders\hrule height .55ex depth -.45ex\hfill\kern0pt} \\
     Evaluated on: & \textsc{EMNIST} $\uparrow$ & \textsc{FMNIST} $\uparrow$ & \textsc{KMNIST} $\uparrow$ & \textsc{EMNIST} $\uparrow$ & \textsc{KMNIST} $\uparrow$ & \textsc{MNIST} $\uparrow$ \\
    \midrule
    \textsc{Map} & $0.877$ & $0.981$ & $0.956$ & $0.684$ & $0.769$ & $0.699$ \\
    \textsc{LA} & $0.638 \pm 0.006$ & $0.745 \pm 0.006$ & $0.681 \pm 0.007$ & $0.432 \pm 0.005$ & $0.436 \pm 0.005$ & $0.433 \pm 0.005$ \\
    \textsc{LinLA} & $0.728 \pm 0.005$ & $0.829 \pm 0.006$ & $0.775 \pm 0.006$ & $0.427 \pm 0.003$ & $0.427 \pm 0.003$ & $0.430 \pm 0.003$ \\
    \textsc{RLA} & $0.846 \pm 0.004$ & $0.939 \pm 0.003$ & $0.882 \pm 0.004$ & $0.556 \pm 0.004$ & $0.586 \pm 0.004$ & $0.574 \pm 0.004$ \\
    \cmidrule(lr){1-7}
    \begin{tabular}[c]{@{}l@{}}\textsc{DiMS} \\ {\footnotesize $t=25.0, \mu=0.1$}\end{tabular} & $0.953 \pm 0.0004$ & $0.979 \pm 0.0004$ & $0.972 \pm 0.0004$ & $0.681 \pm 0.006$ & $0.738 \pm 0.007$ & $0.731 \pm 0.007$ \\
    \begin{tabular}[c]{@{}l@{}}\textsc{DiMS} \\ {\footnotesize $t=50.0, \mu=0.1$}\end{tabular} & $0.961 \pm 0.0002$ & $0.971 \pm 0.0003$ & $0.979 \pm 0.0002$ & $0.715 \pm 0.002$ & $0.779 \pm 0.001$ & $0.781 \pm 0.003$ \\
    \bottomrule
  \end{tabular}%
  }
\end{table}
\begin{table}[ht]
  \centering
  \caption{ID results for a different random seed on MNIST and Fashion MNIST. We show again that the running time does not matter much, compared to the \textsc{RLA} baseline. We also observe that our proposed sampler works well for higher friction coefficients $\eta_0$. We use the low-rank Hessian approximation for all cases, and provide the associated OOD detection results below, given by the \textsc{AUROC} score ($\uparrow$).}
  \label{tab:id-seed1}
  \resizebox{\linewidth}{!}{%
  \begin{tabular}{clccccc}
    \toprule
     &  & \textsc{Acc.} $\uparrow$ & \textsc{NLL} $\downarrow$ & \textsc{Brier} $\downarrow$ & \textsc{ECE} $\downarrow$ & \textsc{MCE} $\downarrow$ \\
    \midrule
    \multirow{6}{*}{\rotatebox{90}{\textsc{MNIST}}} & \textsc{Map} & $0.990$ & $0.034$ & $0.002$ & $0.003$ & $0.125$ \\
     & \textsc{LA} & $0.129 \pm 0.003$ & $62.621 \pm 2.223$ & $0.170 \pm 0.001$ & $0.845 \pm 0.004$ & $0.888 \pm 0.004$ \\
     & \textsc{LinLA} & $0.558 \pm 0.017$ & $4.461 \pm 0.252$ & $0.079 \pm 0.003$ & $0.369 \pm 0.015$ & $0.524 \pm 0.012$ \\
     & \textsc{RLA} & $0.901 \pm 0.005$ & $181.297 \pm 92.880$ & $0.020 \pm 0.001$ & $0.097 \pm 0.005$ & $0.633 \pm 0.010$ \\
    \cmidrule(lr){2-7}
     & \begin{tabular}[c]{@{}l@{}}\textsc{DiMS} \\ {\footnotesize $t=25.0, \eta_0=0.5$}\end{tabular} & $0.967 \pm 0.001$ & $0.161 \pm 0.004$ & $0.005 \pm 0.0001$ & $0.021 \pm 0.0005$ & $0.352 \pm 0.011$ \\
     & \begin{tabular}[c]{@{}l@{}}\textsc{DiMS} \\ {\footnotesize $t=100.0, \eta_0=0.5$}\end{tabular} & $0.945 \pm 0.0003$ & $0.306 \pm 0.002$ & $0.009 \pm 6e-05$ & $0.038 \pm 0.0003$ & $0.367 \pm 0.013$ \\
    \midrule
    \multirow{8}{*}{\rotatebox{90}{\textsc{FMNIST}}} & \textsc{Map} & $0.884$ & $0.316$ & $0.017$ & $0.009$ & $0.737$ \\
     & \textsc{LA} & $0.114 \pm 0.003$ & $87.081 \pm 2.773$ & $0.174 \pm 0.001$ & $0.865 \pm 0.003$ & $0.904 \pm 0.004$ \\
     & \textsc{LinLA} & $0.407 \pm 0.012$ & $5.022 \pm 0.178$ & $0.102 \pm 0.002$ & $0.464 \pm 0.010$ & $0.601 \pm 0.012$ \\
     & \textsc{RLA} & $0.753 \pm 0.001$ & $5.436 \pm 0.098$ & $0.047 \pm 0.0002$ & $0.229 \pm 0.001$ & $0.506 \pm 0.005$ \\
    \cmidrule(lr){2-7}
     & \begin{tabular}[c]{@{}l@{}}\textsc{DiMS} \\ {\footnotesize $t=25.0, \eta_0=0.5$}\end{tabular} & $0.878 \pm 0.001$ & $0.343 \pm 0.003$ & $0.018 \pm 0.0001$ & $0.020 \pm 0.001$ & $0.239 \pm 0.014$ \\
     & \begin{tabular}[c]{@{}l@{}}\textsc{DiMS} \\ {\footnotesize $t=100.0, \eta_0=0.5$}\end{tabular} & $0.879 \pm 0.001$ & $0.383 \pm 0.003$ & $0.018 \pm 9e-05$ & $0.048 \pm 0.001$ & $0.239 \pm 0.011$ \\
    \bottomrule
  \end{tabular}%
  }
  
  \vspace{20pt}
  \resizebox{\linewidth}{!}{%
  \begin{tabular}{lcccccc}
    \toprule
     Trained on: & \multicolumn{3}{c}{\leavevmode\leaders\hrule height .55ex depth -.45ex\hfill\kern0pt\enspace\textsc{MNIST}\enspace\leavevmode\leaders\hrule height .55ex depth -.45ex\hfill\kern0pt} & \multicolumn{3}{c}{\leavevmode\leaders\hrule height .55ex depth -.45ex\hfill\kern0pt\enspace\textsc{FMNIST}\enspace\leavevmode\leaders\hrule height .55ex depth -.45ex\hfill\kern0pt} \\
     Evaluated on: & \textsc{EMNIST} $\uparrow$ & \textsc{FMNIST} $\uparrow$ & \textsc{KMNIST} $\uparrow$ & \textsc{EMNIST} $\uparrow$ & \textsc{KMNIST} $\uparrow$ & \textsc{MNIST} $\uparrow$ \\
    \midrule
    \textsc{Map} & $0.981$ & $0.998$ & $0.991$ & $0.714$ & $0.812$ & $0.751$ \\
    \textsc{LA} & $0.646 \pm 0.008$ & $0.750 \pm 0.008$ & $0.637 \pm 0.008$ & $0.468 \pm 0.006$ & $0.462 \pm 0.006$ & $0.456 \pm 0.007$ \\
    \textsc{LinLA} & $0.741 \pm 0.005$ & $0.885 \pm 0.003$ & $0.767 \pm 0.005$ & $0.390 \pm 0.005$ & $0.409 \pm 0.005$ & $0.375 \pm 0.006$ \\
    \cmidrule(lr){1-7}
    \textsc{RLA} & $0.634 \pm 0.008$ & $0.734 \pm 0.011$ & $0.668 \pm 0.009$ & $0.531 \pm 0.003$ & $0.560 \pm 0.002$ & $0.538 \pm 0.003$ \\
    \begin{tabular}[c]{@{}l@{}}\textsc{DiMS} \\ {\footnotesize $t=100.0, \eta_0=0.5$}\end{tabular} & $0.965 \pm 0.0002$ & $0.975 \pm 0.0005$ & $0.982 \pm 0.0002$ & $0.675 \pm 0.003$ & $0.775 \pm 0.003$ & $0.713 \pm 0.004$ \\
    \begin{tabular}[c]{@{}l@{}}\textsc{DiMS} \\ {\footnotesize $t=25.0, \eta_0=0.5$}\end{tabular} & $0.970 \pm 0.0004$ & $0.988 \pm 0.0003$ & $0.985 \pm 0.0002$ & $0.667 \pm 0.003$ & $0.771 \pm 0.003$ & $0.701 \pm 0.005$ \\
    \bottomrule
  \end{tabular}%
  }
\end{table}
\begin{table}[ht]
  \centering
  \caption{ID results for a third random seed on MNIST and Fashion MNIST. We use the low-rank Hessian approximation for all cases, and provide the associated OOD detection results below, given by the \textsc{AUROC} score ($\uparrow$).}
  \label{tab:id-seed2}
  \resizebox{\linewidth}{!}{%
  \begin{tabular}{clccccc}
    \toprule
     &  & \textsc{Acc.} $\uparrow$ & \textsc{NLL} $\downarrow$ & \textsc{Brier} $\downarrow$ & \textsc{ECE} $\downarrow$ & \textsc{MCE} $\downarrow$ \\
    \midrule
    \multirow{5}{*}{\rotatebox{90}{\textsc{MNIST}}} & \textsc{Map} & $0.989$ & $0.036$ & $0.002$ & $0.004$ & $0.393$ \\
     & \textsc{LA} & $0.136 \pm 0.004$ & $58.633 \pm 2.058$ & $0.169 \pm 0.001$ & $0.838 \pm 0.004$ & $0.885 \pm 0.004$ \\
     & \textsc{LinLA} & $0.583 \pm 0.013$ & $4.013 \pm 0.181$ & $0.074 \pm 0.002$ & $0.347 \pm 0.012$ & $0.508 \pm 0.010$ \\
     & \textsc{RLA} & $0.916 \pm 0.001$ & $10.540 \pm 1.567$ & $0.016 \pm 0.0002$ & $0.081 \pm 0.001$ & $0.585 \pm 0.012$ \\
    \cmidrule(lr){2-7}
     & \begin{tabular}[c]{@{}l@{}}\textsc{DiMS} \\ {\footnotesize $t=50.0, \eta_0=0.5$}\end{tabular} & $0.946 \pm 0.0001$ & $0.291 \pm 0.001$ & $0.009 \pm 3e-05$ & $0.037 \pm 0.0002$ & $0.368 \pm 0.012$ \\
    \midrule
    \multirow{5}{*}{\rotatebox{90}{\textsc{FMNIST}}} & \textsc{Map} & $0.893$ &$0.296$ & $0.015$ & $0.011$ & $0.173$ \\
     & \textsc{LA} & $0.113 \pm 0.004$ & $80.989 \pm 2.723$ & $0.174 \pm 0.001$ & $0.864 \pm 0.004$ & $0.903 \pm 0.004$ \\
     & \textsc{LinLA} & $0.439 \pm 0.014$ & $4.686 \pm 0.209$ & $0.096 \pm 0.002$ & $0.439 \pm 0.012$ & $0.570 \pm 0.012$ \\
     & \textsc{RLA} & $0.755 \pm 0.001$ & $5.850 \pm 0.149$ & $0.047 \pm 0.0002$ & $0.229 \pm 0.001$ & $0.513 \pm 0.005$ \\
    \cmidrule(lr){2-7}
     & \begin{tabular}[c]{@{}l@{}}\textsc{DiMS} \\ {\footnotesize $t=50.0, \eta_0=0.5$}\end{tabular} & $0.880 \pm 0.001$ & $0.349 \pm 0.003$ & $0.018 \pm 0.0001$ & $0.033 \pm 0.001$ & $0.262 \pm 0.016$ \\
    \bottomrule
  \end{tabular}%
  }

    \vspace{20pt}
  \resizebox{\linewidth}{!}{%
  \begin{tabular}{lcccccc}
    \toprule
     & \multicolumn{3}{c}{\leavevmode\leaders\hrule height .55ex depth -.45ex\hfill\kern0pt\enspace\textsc{MNIST}\enspace\leavevmode\leaders\hrule height .55ex depth -.45ex\hfill\kern0pt} & \multicolumn{3}{c}{\leavevmode\leaders\hrule height .55ex depth -.45ex\hfill\kern0pt\enspace\textsc{FMNIST}\enspace\leavevmode\leaders\hrule height .55ex depth -.45ex\hfill\kern0pt} \\
     & \textsc{EMNIST} $\uparrow$ & \textsc{FMNIST} $\uparrow$ & \textsc{KMNIST} $\uparrow$ & \textsc{EMNIST} $\uparrow$ & \textsc{KMNIST} $\uparrow$ & \textsc{MNIST} $\uparrow$ \\
    \midrule
    \textsc{Map} & $0.961$ & $0.987$ & $0.981$ & $0.641$ & $0.758$ & $0.631$ \\
    \textsc{LA} & $0.602 \pm 0.008$ & $0.717 \pm 0.008$ & $0.607 \pm 0.010$ & $0.449 \pm 0.006$ & $0.445 \pm 0.006$ & $0.440 \pm 0.006$ \\
    \textsc{LinLA} & $0.654 \pm 0.006$ & $0.786 \pm 0.006$ & $0.677 \pm 0.006$ & $0.397 \pm 0.007$ & $0.404 \pm 0.007$ & $0.394 \pm 0.008$ \\
    \textsc{RLA} & $0.642 \pm 0.009$ & $0.726 \pm 0.012$ & $0.670 \pm 0.010$ & $0.510 \pm 0.002$ & $0.542 \pm 0.002$ & $0.529 \pm 0.003$ \\
    \cmidrule(lr){1-7}
    \begin{tabular}[c]{@{}l@{}}\textsc{DiMS} \\ {\footnotesize $t=50.0, \eta_0=0.5$}\end{tabular} & $0.965 \pm 0.0003$ & $0.975 \pm 0.001$ & $0.981 \pm 0.0001$ & $0.668 \pm 0.004$ & $0.773 \pm 0.002$ & $0.704 \pm 0.004$ \\
    \bottomrule
  \end{tabular}%
  }
\end{table}

\end{document}